\pdfoutput=1
\PassOptionsToPackage{x11names, dvipsnames, table}{xcolor}
\documentclass[11pt]{article}
\usepackage[T1]{fontenc}
\usepackage[utf8]{inputenc}
\usepackage{amsmath}
\usepackage{amssymb}
\usepackage{mathtools}
\usepackage{adjustbox}
\usepackage{fullpage}
\usepackage{caption}
\usepackage{subcaption}
\usepackage{graphicx}
\usepackage{wrapfig}
\usepackage{cancel}
\usepackage{booktabs}
\usepackage{natbib}
\setcitestyle{authoryear}
\usepackage{url}

\usepackage{amsmath,amsfonts,bm,amsthm}

\def\eqref#1{equation~\ref{#1}}

\def\1{\bm{1}}

\def\rd{{\textnormal{d}}}

\def\va{{\bm{a}}}
\def\vb{{\bm{b}}}

\def\vh{{\bm{h}}}

\def\vm{{\bm{m}}}

\def\vq{{\bm{q}}}

\def\vs{{\bm{s}}}

\def\vu{{\bm{u}}}
\def\vv{{\bm{v}}}
\def\vw{{\bm{w}}}
\def\vx{{\bm{x}}}
\def\vy{{\bm{y}}}
\def\vz{{\bm{z}}}

\def\mI{{\bm{I}}}

\def\mSigma{{\bm{\Sigma}}}

\DeclareMathAlphabet{\mathsfit}{\encodingdefault}{\sfdefault}{m}{sl}
\SetMathAlphabet{\mathsfit}{bold}{\encodingdefault}{\sfdefault}{bx}{n}

\def\gA{{\mathcal{A}}}
\def\gB{{\mathcal{B}}}
\def\gC{{\mathcal{C}}}

\def\gE{{\mathcal{E}}}

\newcommand{\E}{\mathbb{E}}

\newcommand{\R}{\mathbb{R}}

\newcommand{\KL}{D_{\mathrm{KL}}}

\DeclareMathOperator*{\argmin}{arg\,min}

\newtheorem{proposition}{Proposition}
\newtheorem{theorem}{Theorem} %
\theoremstyle{definition} %
\newtheorem{remark}{Remark}
\newtheorem{lemma}{Lemma}
\newtheorem{corollary}{Corollary}

\def\vepsilon{{\bm{\epsilon}}}

\usepackage[dvipsnames]{xcolor}
\usepackage{booktabs}

 \usepackage{bbm}

\DeclareFontFamily{OT1}{pzc}{}
\DeclareFontShape{OT1}{pzc}{m}{it}{<-> s * [1.15] pzcmi7t}{}
\DeclareMathAlphabet{\mathpzc}{OT1}{pzc}{m}{it}

\usepackage{tikz}
\usepackage[scaled=0.95]{helvet}
\usetikzlibrary{positioning, arrows.meta, calc, backgrounds, fit}
 
\definecolor{cSample}{HTML}{3B82C4}
\definecolor{cRoll}{HTML}{2A9D8F}
\definecolor{cWeight}{HTML}{B5651D}
\definecolor{cUpdate}{HTML}{7C5CBF}
\definecolor{cIO}{HTML}{6B7280}
\definecolor{cBG}{HTML}{F7F8FA}

\definecolor{RevisionNavy}{HTML}{1B3F8B}
\colorlet{ArxivBlue}{RevisionNavy!50!white} %
\newif\ifshowarxiv
\showarxivfalse
\newcommand{\revisionarxiv}[1]{\ifshowarxiv{\color{ArxivBlue}#1}\else\ignorespaces\fi}
\usepackage{environ} %
\NewEnviron{revisionarxivblock}{\ifshowarxiv\par{\color{ArxivBlue}\BODY}\par\fi}

\tikzset{
  blk/.style 2 args={
    rectangle, rounded corners=7pt, draw=#1, line width=1.1pt,
    fill=#1!12, inner sep=7pt, align=center,
    minimum height=3.2cm, text width=#2},
  ioblk/.style={rectangle, rounded corners=5pt, draw=cIO, line width=0.9pt,
    fill=white, inner sep=5pt, align=center, font=\footnotesize},
  flow/.style={-{Stealth[length=3mm, width=2.2mm]}, line width=1.2pt, draw=#1},
  flowlab/.style={font=\fontfamily{phv}\selectfont\bfseries\scriptsize, text=#1!55!black, fill=cBG, inner sep=2pt},
}

\usepackage[most]{tcolorbox}

\definecolor{propositionblue}{RGB}{246,248,253}

\tcbset{statementbox/.style={
    enhanced,
    breakable,
    colback=propositionblue,
    colframe=propositionblue,
    boxrule=0pt,
    arc=0pt,
    outer arc=0pt,
    boxsep=0pt,
    left=1.4mm,
    right=1.4mm,
    top=1.4mm,
    bottom=1.4mm,
    before skip=5pt plus 1pt minus 1pt,
    after skip=5pt plus 1pt minus 1pt
}}

\tcolorboxenvironment{proposition}{statementbox}
\tcolorboxenvironment{theorem}{statementbox}
\tcolorboxenvironment{lemma}{statementbox}
\tcolorboxenvironment{corollary}{statementbox}

\makeatletter
\newcommand{\statementbox@safelabel}{%
  \let\statementbox@label\label
  \def\label##1{%
    \ifvmode
      \everypar\expandafter{\the\everypar\statementbox@label{##1}}%
    \else
      \statementbox@label{##1}%
    \fi}}
\AtBeginEnvironment{proposition}{\statementbox@safelabel}
\AtBeginEnvironment{theorem}{\statementbox@safelabel}
\AtBeginEnvironment{lemma}{\statementbox@safelabel}
\AtBeginEnvironment{corollary}{\statementbox@safelabel}
\makeatother

\definecolor{VennControl}{HTML}{E8A33D}
\definecolor{VennPolicy}{HTML}{B3261E}
\definecolor{VennLTO}{HTML}{7655C9}
\definecolor{VennNavy}{HTML}{173B67}
\definecolor{VennText}{HTML}{142638}
\colorlet{PanelText}{VennText}
\definecolor{PlaceholderFill}{HTML}{F7F9FB}
\definecolor{PlaceholderLine}{HTML}{AAB6C2}
\definecolor{PlaceholderText}{HTML}{748392}

\definecolor{FlowBlue}{RGB}{45, 120, 190}       %
\definecolor{FlowFigBlue}{HTML}{214DCF}          %
\definecolor{MPPICoral}{RGB}{225, 90, 60}       %
\definecolor{CEMGold}{RGB}{225, 170, 40}        %
\definecolor{CMATeal}{RGB}{40, 160, 130}        %
\definecolor{DIALIndigo}{RGB}{60, 40, 140}      %

\usepackage[scaled=1.0]{inconsolata}
\usepackage{xspace}
\newcommand{\flowoptpolicy}{%
\textcolor{FlowBlue}{%
\texttt{\fontseries{b}\selectfont FlowOpt\hspace{0.3em}Policy}%
}\xspace}

\newcommand{\mppi}{\textcolor{MPPICoral}{\texttt{MPPI}}}
\newcommand{\cem}{\textcolor{CEMGold}{\texttt{CEM}}}
\newcommand{\cmaes}{\textcolor{CMATeal}{\texttt{CMA-ES}}}
\newcommand{\dialmpc}{\textcolor{DIALIndigo}{\texttt{DIAL-MPC}}}
\definecolor{LGURed}{RGB}{139, 26, 26}
\newcommand{\lgu}{\textcolor{LGURed}{\texttt{LGU}}}                                 %
\newcommand{\lgufull}{\textcolor{LGURed}{\texttt{Learned Gaussian Update (LGU)}}}   %

\definecolor{FlowPolicySky}{RGB}{135, 175, 215}   %
\definecolor{FlowPolicyPink}{HTML}{C0399B}
\newcommand{\flowpolicy}{\textcolor{FlowPolicyPink}{\texttt{Flow\hspace{0.3em}Policy}}}

\usepackage{enumitem}

\usepackage{hyperref}
\hypersetup{
    colorlinks=true,
    linkcolor=blue,
    filecolor=magenta,
    urlcolor=blue,
    citecolor=blue,
}

\newif\ifsansheadings
\sansheadingsfalse
\newcommand{\headingsf}{\ifsansheadings\fontfamily{cmss}\selectfont\fi}
\newcommand{\titlefont}{\normalfont\LARGE\serifbfseries\boldmath}

\usepackage[skip=6pt plus 1pt]{parskip}

\usepackage{abstract}
\usepackage{titlesec}
\def\sectionfont{\headingsf\Large\bfseries\boldmath}
\def\subsectionfont{\headingsf\large\bfseries\boldmath}
\def\paragraphfont{\headingsf\normalsize\bfseries\boldmath}
\titleformat*{\section}{\sectionfont}
\titleformat*{\subsection}{\subsectionfont}
\titleformat*{\subsubsection}{\paragraphfont}
\titleformat*{\paragraph}{\paragraphfont}
\titleformat*{\subparagraph}{\paragraphfont}

\makeatletter
\let\serifbfseries\bfseries
\def\arxiv@hsf{cmss}
\renewcommand{\bfseries}{%
  \edef\arxiv@rm{\rmdefault}%
  \ifsansheadings \ifx\f@family\arxiv@rm \fontfamily{cmss}\fi \fi
  \ifx\f@family\arxiv@hsf
    \fontseries{bx}\selectfont
  \else
    \serifbfseries
  \fi}
\makeatother
\DeclareCaptionFont{headingsf}{\headingsf}
\makeatletter
\renewcommand{\@maketitle}{
    \begin{center}
        {\titlefont \@title \par}
        \vskip 1.5em
        {\large \@author \par}
        \vskip 1em
        {\large \@date \par}
    \end{center}
}
\makeatother

\makeatletter
\def\th@plain{\normalfont}
\makeatother

\colorlet{RevisionNavy}{black}
\colorlet{NavyBlue}{black}
\colorlet{ArxivBlue}{black}
\showarxivtrue

\makeatletter
\let\arxiv@vspace\vspace
\newcommand{\arxiv@vsp}[2]{%
  \setlength\@tempskipa{#2}%
  \ifdim\@tempskipa<\z@ \else \arxiv@vspace#1{#2}\fi}
\renewcommand{\vspace}{\@ifstar{\arxiv@vsp{*}}{\arxiv@vsp{}}}
\makeatother

\title{FlowOpt Policy: \\ Learning to Optimize Actions with Flows}

\author{
    \large
    Augustinos D. Saravanos\addtocounter{footnote}{1}\thanks{Corresponding author: \texttt{asaravan@mit.edu}}\hphantom{\textsuperscript{\textdagger}}, Oswin So and Chuchu Fan
    \\[0.5em]
    Massachusetts Institute of Technology
}

\date{}

\begin{document}
\maketitle

\begin{abstract}
Model-based planning requires repeatedly solving complex non-convex optimization problems over action sequences. Sampling-based optimizers are most commonly used, yet their handcrafted updates cannot exploit structure shared across planning instances. From a complementary perspective, expressive generative policies produce action sequences directly without optimizing them over the planning horizon using a predictive model.
We bridge these two paradigms by introducing \textbf{FlowOpt Policy}, which casts the optimization of action sequences as a learned transport problem.
Rather than reweighting candidates and resampling from a proposal, or
generating actions directly, a learned flow iteratively transports the candidate population based on its rollout evaluations.
Our framework is theoretically grounded in a splitting of the Wasserstein gradient flow of a free energy over action populations.
The resulting proximal step defines flow-matching targets using candidates obtained by a higher-budget optimizer from the populations FlowOpt Policy encounters, enabling self-supervised training without demonstration data.
The same formulation also motivates attention over the population, allowing each candidate's update to depend on the others' locations and rollout costs.
We further show that classical methods such as MPPI and CEM emerge as special cases of this mechanism.
Across physics-based and learned world-model tasks, FlowOpt Policy outperforms both handcrafted and learned sampling-based planners, and generalizes better than state-conditioned generative policies to unseen objectives, as it learns how to optimize rather than what to output.

\end{abstract}

\section{Introduction}

An autonomous agent equipped with a predictive model of its environment has the ability to plan ahead, since it can select actions by optimizing their simulated consequences before executing them \citep{betts2010practical, moerland2023model}. 
This paradigm is central to sequential decision-making across AI, with applications ranging from robotics \citep{lavalle2006planning} to autonomous navigation \citep{paden2016survey} and interactive agents acting in simulated or virtual environments \citep{schrittwieser2020mastering}. 
The underlying predictive model may range from physics simulators \citep{todorov2012mujoco, makoviychuk2021isaac} to learned world models that predict future outcomes in latent space from raw observations \citep{ha2018world, hansen2024tdmpc2, zhou2024dino, assran2025v}. 
Regardless of the model's form, the induced optimization problems over action sequences are most often highly non-convex, with actions tightly coupled across the horizon through the dynamics. 
Moreover, when deployed in closed loop as model predictive control (MPC), these problems must be re-solved under strict computational budgets, which makes their efficient solution critical.

\ifshowarxiv\newpage\fi
While gradient-based methods excel in smooth, well-behaved landscapes \citep{jacobson1970differential, li2004iterative}, sampling-based planners such as Cross-Entropy Method (CEM) \citep{deboer2005cem_tutorial} and Model Predictive Path Integral (MPPI) \citep{williams2017information}, have become the method of choice for planning under non-smooth, contact-rich, and learned models, thanks to their derivative-free and highly parallelizable nature \citep{howell2022predictive, bhardwaj2022storm}. 
Nevertheless, these planners rely on fixed, handcrafted update rules, and their performance hinges on carefully tuned hyperparameters and large sample budgets.
\ifshowarxiv Their\else More fundamentally, their\fi{} core optimization mechanism compresses the population into the parameters of a proposal distribution through a fixed update rule, and then draws new candidates from that distribution.
The same rule is moreover applied to every problem instance, so patterns shared across problems from the same task distribution are never exploited.
This raises a natural question: can the transport of the population itself be learned, so that candidates move in a problem-adapted way given how they scored?

\begin{figure}[t]
\vspace{-0.2cm}
    \centering
    \resizebox{\textwidth}{!}{%
        \begin{tikzpicture}[
            x=1cm,
            y=1cm,
            font=\sffamily,
            flowarrow/.style={
                -{Latex[length=2.8mm,width=2mm]},
                line width=1.25pt,
                draw=FlowFigBlue
            },
            plotaxis/.style={
                -{Latex[length=1.8mm]},
                line width=0.75pt,
                draw=VennText
            }
        ]
 
        \path[use as bounding box] (0,0) rectangle (17.5,5.3);
 
        \fill[PlaceholderFill,rounded corners=4pt]
            (0.10,0.25) rectangle (12.55,5.01);
        \draw[PlaceholderLine,draw opacity=0.35,
              line width=0.8pt,rounded corners=4pt]
            (0.10,0.25) rectangle (12.55,5.01);
 
        \node[align=center,anchor=north,font=\sffamily\bfseries\small,
              text=VennText,inner sep=0pt]
            at (1.80,4.75) {Initial action\\sequence population};
        \node[align=center,anchor=north,font=\sffamily\bfseries\small,
              text=VennText,inner sep=0pt]
            at (10.85,4.75) {Final action\\sequence population};
 
        \foreach \cx/\img in {%
            1.80/{\includegraphics[width=2.90cm]{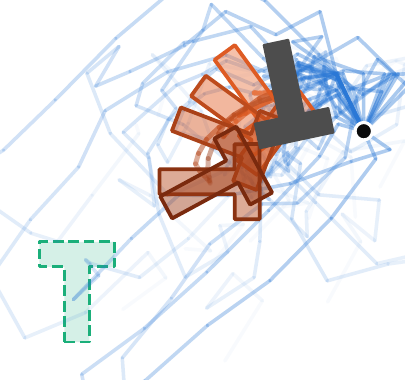}},%
            10.85/{\includegraphics[width=2.90cm]{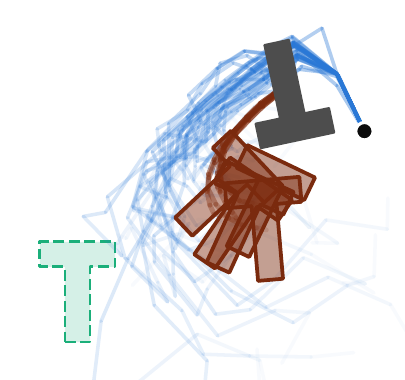}}} {
            \node[inner sep=0pt] (img) at (\cx,2.58)
                {\img};
            \draw[black,line width=0.7pt]
                (img.south west) rectangle (img.north east);
            \node[inner sep=0pt,rotate=-90] (cbar) at (\cx,0.96)
                {\includegraphics[width=0.13cm,height=2.90cm]
                    {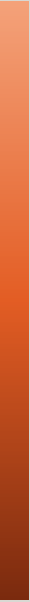}};
            \draw[black,line width=0.4pt]
                (cbar.south west) rectangle (cbar.north east);
            \node[anchor=base west,font=\sffamily\scriptsize,text=PanelText]
                at (\cx-1.45,0.60) {low};
            \node[anchor=base east,font=\sffamily\scriptsize,text=PanelText]
                at (\cx+1.45,0.60) {high};
            \node[anchor=base,font=\sffamily\scriptsize\bfseries,text=PanelText]
                at (\cx,0.60) {plan cost};
        }
 
        \draw[flowarrow] (3.32,2.58)--(5.22,2.58);
        \node[align=center,anchor=south,font=\sffamily\scriptsize,
              text=VennText,inner sep=1pt]
            at (4.27,2.67) {candidates \&\\rollout costs};
 
        \draw[fill=white,draw=FlowFigBlue,draw opacity=0.70,
              line width=1.05pt,rounded corners=3pt]
            (5.29,0.73) rectangle (8.52,4.63);
        \node[font=\sffamily\bfseries\normalsize,text=FlowFigBlue]
            at (6.90,4.30) {FlowOpt Policy};
        \node[font=\sffamily\small,text=VennText]
            at (6.90,3.93) {population transport};
 
        \draw[plotaxis] (5.85,1.53)--(8.25,1.53);
        \draw[plotaxis] (5.85,1.53)--(5.85,3.73);
        \node[rotate=90,font=\sffamily\scriptsize,text=VennText]
            at (5.55,2.63) {best cost};
        \node[font=\sffamily\scriptsize,text=VennText]
            at (5.86,1.32) {$0$};
        \node[font=\sffamily\scriptsize,text=VennText]
            at (8.00,1.32) {$K$};
        \node[font=\sffamily\scriptsize,text=VennText]
            at (7.00,0.96) {optimization round};
        \draw[FlowFigBlue,line width=1.65pt,line cap=round,
              rounded corners=3pt]
            (5.99,3.50) -- (6.30,3.43) -- (6.53,2.95)
            -- (6.81,2.58) -- (7.16,2.26) -- (7.51,2.09)
            -- (8.03,1.98);
        \fill[FlowFigBlue] (5.99,3.50) circle (1.5pt);
        \fill[FlowFigBlue] (8.03,1.98) circle (1.5pt);
 
        \draw[flowarrow] (8.58,2.58)--(9.34,2.58);
 
        \node[draw=VennControl,fill=VennControl!12,line width=0.9pt,
              rounded corners=3pt,align=center,inner xsep=5pt,inner ysep=4pt,
              font=\sffamily\scriptsize\bfseries,text=VennControl!70!black,
              minimum width=1.25cm] (predmodel) at (4.27,1.48)
            {Predictive\\model};
        \draw[flowarrow,rounded corners=3pt]
            (6.90,0.73) -- (6.90,0.51) -- (4.27,0.51) -- (predmodel.south);
        \draw[FlowFigBlue,line width=1.25pt] (predmodel.north) -- (4.27,2.58);
 
        \begin{scope}[
            shift={(15.30,2.72)},
            scale=0.475,
            transform shape
        ]
 
        \coordinate (policy)  at (0,0.85);
        \coordinate (control) at (-1.22,-0.95);
        \coordinate (lto)     at (1.22,-0.72);
 
        \draw[VennPolicy,draw opacity=0.10,line width=7pt]
            (policy) circle[radius=2.85];
        \draw[VennControl,draw opacity=0.10,line width=7pt]
            (control) circle[radius=2.85];
        \draw[VennLTO,draw opacity=0.10,line width=7pt]
            (lto) circle[radius=2.85];
 
        \filldraw[fill=VennPolicy,fill opacity=0.17,
                  draw=VennPolicy,draw opacity=0.92,line width=1.35pt]
            (policy) circle[radius=2.85];
        \filldraw[fill=VennControl,fill opacity=0.17,
                  draw=VennControl,draw opacity=0.92,line width=1.35pt]
            (control) circle[radius=2.85];
        \filldraw[fill=VennLTO,fill opacity=0.17,
                  draw=VennLTO,draw opacity=0.92,line width=1.35pt]
            (lto) circle[radius=2.85];
 
        \node[align=center,font=\sffamily\bfseries\Large,
              text=VennPolicy,text width=5.0cm]
            at (0,4.4) {Generative Policies};
        \node[align=center,font=\sffamily\bfseries\Large,
              text=VennControl!80!black,text width=5.0cm]
            at (-2.25,-4.70) {Model Predictive \\ Control};
        \node[align=center,font=\sffamily\bfseries\Large,
              text=VennLTO,text width=5.0cm]
            at (2.25,-4.70) {Learning-to-\\Optimize};
 
        \node[rounded corners=2.5pt,fill=FlowFigBlue,draw=white,
              line width=0.85pt,inner xsep=12pt,inner ysep=7pt,
              align=center,text=white,
              font=\sffamily\bfseries\Large]
            at (0,-0.55) {FlowOpt\\Policy};
 
        \end{scope}
        \end{tikzpicture}%
    }
\vspace{-0.8cm}
\caption{\textbf{Conceptual overview.} Our method \textbf{FlowOpt Policy} iteratively transports a population of candidate action sequences via a learned flow\ifshowarxiv\else\space model\fi{} toward lower cost, conditioned on their rollout scores,\ifshowarxiv\else\space essentially\fi{} bridging model predictive control, generative modeling and learning-to-optimize.}
\label{fig: intro figure}
\vspace{-0.5cm}

\end{figure}

At the other end of the spectrum, generative policies based on diffusion and flow models capture complex distributions over action sequences conditioned on states/observations and, when available, additional task context such as a goal \citep{ajay2022isconditional, janner2022diffuser, chi2025diffusionpolicy, black2024pi_0}.
Such policies, however, use this information to directly generate an action sequence, without evaluating candidates through a predictive model and querying their rollout costs to iteratively refine them at deployment.
As a result, their performance depends on the training distribution, and changes in the cost or constraints at deployment may require retraining, potentially limiting their ability to generalize well across deployment settings.

These two paradigms thus exhibit complementary limitations: sampling-based planners optimize at deployment but do not amortize across instances, while generative policies learn from data but perform no optimization at deployment. 
This suggests a third option: learn the planner's optimization procedure itself, so that deployment retains an iterative, model-in-the-loop search across action sequences while its update rule is shaped by data.
This viewpoint connects with another field, learning to optimize (L2O), where an optimizer is trained on a distribution of problems and then applied to new instances in place of a hand-designed method \citep{chen2022learning, amos2023tutorial}.
Sampling-based planners invite this treatment, yet existing learned variants remain limited, primarily learning the parameters of the sampling distribution \citep{sacks2022learning, power2022variational}.

\paragraph{Contributions.} Our key insight is that a sampling-based planner's iteration can be viewed as a map transporting a population of candidate action sequences, conditioned on their evaluations under the model, precisely the class of objects that modern generative models represent. Building on this perspective, we introduce \textbf{FlowOpt Policy}, a learned flow that transports action populations, conditioned only on their rollout evaluations (Fig. \ref{fig: intro figure}). Since it is conditioned on costs rather than states, FlowOpt Policy can optimize unseen objectives without retraining. Training is self-supervised and demonstration-free, using flow matching to learn from improvements computed by a sampling-based optimizer with a larger budget. FlowOpt Policy substantially outperforms sampling-based planners, and, unlike state-conditioned generative policies, generalizes to unseen objectives, both in physics-based simulators and in learned world models.
Our main contributions are the following:
\begin{itemize}
\item We introduce a novel self-supervised framework that casts planning as a learned transport of a population of candidate action sequences. Each planning iteration is represented by a conditional flow that improves the candidates using the population and its rollout costs.
\item We theoretically ground our framework in a heat--proximal splitting of the Wasserstein gradient flow of a free energy, where the heat substep re-expands the population through perturbations, while the proximal substep yields principled training targets. 
\item We show that attention arises naturally from the flow-matching and proximal formulations. In our parametrization, each candidate attends to other members through their costs, 
yielding an update that is invariant to the scale of actions and costs, which allows it to generalize across problem instances. We further show that the updates of classical sampling-based planners such as MPPI and CEM correspond to fixed attention patterns.
\item We provide an experimental demonstration on physics-based and world model tasks, showing that our approach consistently outperforms well-established sampling-based planners, and exhibits substantially better generalization than state-conditioned generative policies.
\end{itemize}

\section{Related Work}
\vspace{-0.1cm}

\paragraph{Control as inference.}
Optimal control admits a well-established inference formulation: KL and path-integral control express the optimal trajectory distribution as an exponential reweighting of a prior by the trajectory cost, arising from free-energy minimization \citep{kappen2012optimal, theodorou2012relative, rawlik2012stochastic}.
In reinforcement learning, the same structure underlies maximum-entropy RL and RL as inference, where optimal policies arise by conditioning a reference policy on optimality \citep{levine2018reinforcement, haarnoja2018soft, eysenbach2021maximum}. 
Many sampling-based planners rest on the same perspective, approximating Gibbs distributions over action sequences through different prescribed update rules \citep{williams2017information, okada2020variational}.
Our method shares this perspective, but learns the optimization procedure itself, as a transport of the population of candidate action sequences, rather than prescribing the update rule, while grounded in a Wasserstein proximal step derived from the same free-energy objective.
\vspace{-0.2cm}
\paragraph{Sampling-based predictive control.}
Methods such as MPPI \citep{williams2017information}, CEM \citep{deboer2005cem_tutorial}, CMA-ES \citep{hansen2001cmaes} etc., optimize action sequences by sampling candidates from a parametric proposal, scoring them through model rollouts, and refitting the proposal around the best-scoring ones. 
Model-based diffusion \citep{pan2024modelbaseddiffusion} estimates the score of a noise-smoothed cost from rollouts, making each reverse step an annealed MPPI-style average, while DIAL-MPC \citep{xue2025dial} anneals the sampling variance across iterations and the horizon.
OT-MPC \citep{pacelli2026sampling} refines candidates toward promising samples through a fixed entropy-regularized optimal transport coupling.
Amortization has mainly entered these planners through the proposal.
\citet{sacks2023learning} and \citet{power2022variational} learn richer sampling distributions, while TD-MPC and TD-MPC2 \citep{hansen2022temporal, hansen2024tdmpc2} seed the candidates with a learned policy prior. In all these methods, the update remains the fixed, classical cost-weighted rule.

\vspace{-0.2cm}
\paragraph{Generative action policies.}
Diffusion and flow models have become a standard policy class for control, generating action sequences conditioned on observations \citep{chi2025diffusionpolicy} or on returns and constraints \citep{ajay2022isconditional}. A second line generates full trajectories and steers the sampling process at test time, through value-gradient guidance \citep{janner2022diffuser}, constraint enforcement inside the reverse process \citep{xiao2025safediffuser, zhang2026constrained}, or reward-guided self-improvement \citep{liang2023adaptdiffuser}. Closest to us are methods that pair a generator with a predictive model: D-MPC \citep{zhou2025diffusionmpc} learns a diffusion action proposal alongside a diffusion dynamics model and plans by sampling, scoring and ranking, while \cite{kurtz2025generative} train a state-conditioned flow policy on the outputs of a sampling-based planner and use its samples to warm-start it.

\vspace{-0.2cm}
\paragraph{Learning-to-optimize.}
Learning-to-optimize (L2O) replaces hand-designed optimization algorithms with learned ones trained on a distribution of problem instances \citep{chen2022learning, amos2023tutorial}. More recently, generative models have been repurposed as optimizers that sample from low-cost or feasible regions \citep{krishnamoorthy2023diffusion, kong2024diffusion, uehara2024bridging, saravanos2026learningtooptimize_via_flows}. 
Within trajectory optimization and MPC, L2O approaches have mostly been used to warm-start solvers and accelerate optimizers \citep{sambharya2023end, saravanos2025deep} or through diffusion and flow models that propose candidate trajectories then refined by a solver \citep{li2024diffusolve}. \citet{sacks2022learning, sacks2024deep} learn to update the parameters of a Gaussian proposal that redraws candidates, whereas we learn the transport of the candidates themselves.

\vspace{-0.1cm}
\section{Background}
\label{sec:background}
\vspace{-0.1cm}
\subsection{Problem Setting}
\vspace{-0.1cm}
Model predictive control (MPC) is a framework for sequential decision-making in which a finite-horizon sequence of actions is optimized using a predictive model. 
At its core, the underlying optimization problem that is being solved recursively in a receding horizon fashion has the form: 
\begin{equation}
\min_{\bar{\va}} ~ J(\vs_0, \bar{\va}) 
:= \sum_{t=0}^{H-1} \ell(\vs_t, \va_t) + \phi (\vs_H)  
\quad
\mathrm{s.t.} \quad \vs_{t+1} = f(\vs_t, \va_t), \quad \vs_0 = \hat{\vs}_0, 
\label{eq: opt_problem}
\end{equation}
where $\vs_t \in \R^{n_s}$ and $\va_t \in \R^{n_a}$ denote the state and action at time step $t$, $\hat{\vs}_0$ is a given initial state, $f(\cdot)$ describes the system dynamics and $J(\cdot)$ defines the cost to minimize (or reward to maximize) consisting of a running term $\ell(\cdot)$ and a terminal one $\phi(\cdot)$. The state and action sequences over a planning horizon $H$ are defined as
$\bar{\vs} = \{\vs_0, \dots, \vs_H\}$ and 
$\bar{\va} = \{\va_0, \dots, \va_{H-1}\}$, respectively.
For notational convenience, we will treat $J$ as a rollout oracle without explicitly writing the dynamics.
\vspace{-0.4cm}
\subsection{Sampling-Based Optimization}
\vspace{-0.1cm}
Sampling-based methods such as MPPI \citep{williams2017information} and CEM \citep{deboer2005cem_tutorial}, among others, address problem (\ref{eq: opt_problem}) by iteratively sampling from a parametric distribution, evaluating candidate action sequences and updating the distribution parameters.
In particular, at optimization iteration $k$, a population of $P$ candidate action sequences $\smash[b]{\{\bar{\va}_{j}^{k}\}_{j=1}^P}$ is drawn around the current nominal sequence $\bar{\va}_{\mathrm{nom}}^{k}$ by adding Gaussian perturbations $\vepsilon_{j}^{k}$ as follows:
\begin{equation}
\bar{\va}_{j}^{k} = \bar{\va}_{\mathrm{nom}}^{k} + \vepsilon_{j}^{k}, \qquad \vepsilon_{j}^{k} \sim \mathcal{N}(\mathbf{0}, \mSigma), \qquad j = 1,\dots,P,
\label{eq: sampl_opt candidates}
\end{equation}
and each candidate is scored by $J_j^{k} := J(\hat{\vs}_0, \bar{\va}_{j}^{k})$, which entails rolling out through the dynamics (\ref{eq: opt_problem}).
The nominal sequence is then updated as a cost-weighted average of the sampled perturbations:
\vspace{-0.1cm}
\begin{equation}
\bar{\va}_{\mathrm{nom}}^{k+1} =
\bar{\va}_{\mathrm{nom}}^{k} + \sum_{j=1}^{P} w_j^{k} \vepsilon_j^{k},
\qquad
w_j^{k} = \frac{g \big(J_j^{k} \big)}
{\sum_{i=1}^{P} g \big(J_i^{k} \big)}.
\label{eq: sampl_opt update}
\end{equation}
Different sampling-based algorithms correspond to different choices of the weighting function $g(\cdot)$.
In MPPI, the weights are defined through exponential cost shaping, 
$g_{\mathrm{MPPI}}(J_j) = \exp\left(-J_j / \lambda \right)$,
where $\lambda>0$ is the temperature parameter. 
In CEM, only an elite set $\mathcal{E}$ of the lowest-cost samples is retained,
$g_{\mathrm{CEM}}(J_j) = \mathbf{1}\!\left\{j \in \mathcal{E}\right\}$, so that elites are weighted uniformly and the rest are discarded.
\vspace{-0.1cm}
\section{Learning to Optimize Actions via Flows}

\vspace{-0.2cm}
\subsection{Planning as Transport of Action Populations}
\label{sec: method-transport}
\vspace{-0.2cm}
Sampling-based methods evolve populations of action sequences by scoring them, compressing them into a parametric proposal distribution and resampling.
These methods differ mainly in their hand-designed rules for collapse and re-expansion.
We instead view planning as an iterative map from a population and its scores to a new population.
This motivates framing optimization for planning as a transport over populations of action sequences, where we learn how candidates should move based on their scores.
Let a planner maintain a population of \(P\) candidate action sequences:
\begin{equation}
\gA^k = \{\bar{\va}_j^k\}_{j=1}^P, \qquad \bar{\va}_j^k = \big\{ \va_{j,0}^k, \dots, \va_{j,H-1}^k \big\} \in \R^{H \times n_a},
\label{eq:population}
\end{equation}
at optimization iteration $k$. Given the current state $\hat{\vs}_0$, each candidate $\bar{\va}_j^k$ is rolled out under the dynamics (\ref{eq: opt_problem}), to obtain its cost $\smash{J_j^k := J(\hat{\vs}_0, \bar{\va}_j^k)}$.
All information available to the planner at iteration $k$ (i.e., the population and its costs) is collected in the \emph{iteration context} $\gC^k$ specified below.
One planning iteration then corresponds to a transport of the population conditioned on the context,
\begin{equation}
\gA^{k+1} = \Phi_\theta \big(\gA^k;\, \gC^k\big),
\label{eq:transport}
\end{equation}
where $\theta$ are learnable parameters. 
Our goal is to learn a transport $\Phi_\theta$ that exploits the rollout costs directly, with population recombination determined by data rather than prescribed weighting rules.

This also distinguishes our approach from standard generative control policies, which typically learn to generate action sequences conditioned on states/observations and additional task context, i.e., $\bar{\va} \sim \pi_\theta(\cdot|\hat{\vs}_0, \mathrm{Context})$, without optimizing them through a predictive model at deployment. The objective they were trained on is absorbed into $\theta$, and their performance fundamentally relies on the training distribution capturing every state and objective met at deployment, with no mechanism to correct a generated sequence once its cost is known. Our framework instead generates
\begin{equation}
\mathcal{A}^{K} = \Phi_\theta\big(\,\cdot\,;\mathcal{C}^{K-1}\big) \circ \cdots \circ \Phi_\theta\big(\,\cdot\,;\mathcal{C}^{0}\big)\big(\mathcal{A}^{0}\big), \qquad
\bar{\va}^{\star} = \argmin_{\bar{\va} \in \mathcal{A}^{K}} J(\hat{\vs}_0, \bar{\va}),
\label{eq:planner_output}
\end{equation}
so that the objective enters at deployment through the contexts $\mathcal{C}^k$ rather than through the weights, and essentially learns how to use the costs to improve the population.
\vspace{-0.2cm}
\paragraph{Lifting to measures and free-energy formulation.}
\ifshowarxiv We\else Let us\fi{} lift (\ref{eq: opt_problem}) from sequences to probability measures $\nu$ over $\R^{d}$, $d = H n_a$, and consider the free-energy formulation \citep{jordan1998variational}:
\begin{equation}
\min_{\nu}\ F_\lambda(\nu) := \E_\nu[J] + \lambda\, \mathcal{H}(\nu), \qquad \mathcal{H}(\nu) := \int \rho \log \rho \, \rd\bar{\va} \quad \text{for } \rd\nu = \rho\, \rd\bar{\va}.
\label{eq: free energy}
\end{equation}
The first term, $\E_\nu[J]$, is the expected cost of a sequence drawn from $\nu$. Its minimum over $\nu$ equals the optimal value of (\ref{eq: opt_problem}).
The second term, $\lambda\,\mathcal{H}(\nu)$, scales negative entropy by the temperature $\lambda>0$ to maintain population spread. We set $\mathcal{H}(\nu)=+\infty$ when $\nu$ has no density.
The minimizer of (\ref{eq: free energy}) is the Gibbs measure $\smash{\nu_\lambda \propto e^{-J/\lambda}}$, concentrating on the minimizers of $J$ as $\lambda \to 0$ (Appendix~\ref{app: theory - lift}). 
Most sampling-based planners are related to this variational formulation: MPPI \citep{williams2017information} takes the entropy relative to a Gaussian proposal around the nominal sequence, CEM \citep{deboer2005cem_tutorial} instead performs a KL projection onto an elite-truncated proposal distribution, and diffusion-style planners \citep{pan2024modelbaseddiffusion,xue2025dial} progressively anneal their sampling noise.
\vspace{-0.2cm}
\paragraph{Wasserstein proximal step and JKO scheme.} 
Reweighting evaluated candidates leaves their support unchanged, motivating a formulation that transports candidates directly in action space.
We therefore view planning as a Wasserstein proximal step on $F_\lambda$, which penalizes population displacement rather than merely reweighting candidates.
Let $\vx = (\vx_1, \dots, \vx_P) \in \R^{P \times d}$ denote the stacked population, with $\vx_j = \operatorname{vec}(\bar{\va}_j)$, identified with the measure $\smash{\mu_{\vx} = \frac{1}{P}\sum_{j} \delta_{\vx_j}}$.
The Wasserstein distance between populations is their mean squared displacement minimized over one-to-one matchings $\sigma$, $\smash{W_2^2(\mu_{\vx}, \mu_{\vy}) = \min_{\sigma} \frac{1}{P}\sum_{j} \|\vx_j - \vy_{\sigma(j)}\|^2}$\revisionarxiv{~(general measures in Appendix~\ref{app: jko step})}.
The proximal step of $F_\lambda$ with step size $h > 0$ then yields the JKO step \citep{jordan1998variational},
\begin{equation}
\nu^{k+1} \in \argmin_{\nu}\ F_\lambda(\nu) + \frac{1}{2h}\, W_2^2(\nu, \nu^k).
\label{eq: jko}
\end{equation}
\vspace{-0.7cm}
\paragraph{Heat--proximal splitting.} Computing the JKO step (\ref{eq: jko}) directly requires more than a finite set of evaluated candidates, since the entropy forces its minimizer to have a density (Lemma~\ref{lem: jko}), and the expected cost requires knowing $J$ wherever mass may be sent. We therefore split the JKO step into two substeps of size $h$ (entropy first, expected cost second), which yields
\begin{equation}
\nu^{k+\frac{1}{2}} = \nu^k * \mathcal{N}(0, 2\lambda h I), \qquad \nu^{k+1} \in \argmin_{\nu}\ \E_\nu[J] + \frac{1}{2h}\, W_2^2(\nu, \nu^{k+\frac{1}{2}}).
\label{eq: splitting}
\end{equation}
The first substep is the Wasserstein gradient flow of the entropy over a step $h$, i.e., the heat flow, available in closed form.
The second substep is the Wasserstein proximal step of the expected cost, taken implicitly since the cost flow generally has no closed-form solution.
This is a heat--proximal splitting \citep{bernton2018langevin}, with one exact and one proximal substep as in implicit--explicit schemes.

\begin{proposition}[Heat--proximal splitting of the JKO step]\label{prop: splitting}
Let $J$ be lower semicontinuous and bounded below. Then, the two substeps in (\ref{eq: splitting}) admit the following characterizations.
\begin{enumerate}[label=(\roman*), leftmargin=2.2em, itemsep=0pt, topsep=2pt]
\item \textbf{Heat substep.} The first substep is the Wasserstein gradient flow of $\lambda \mathcal{H}$ over a step $h$, which is sampled by perturbing the population, $\smash{\vx^0_j = \vx_j + \bm{\xi}_j}$ with i.i.d.\ $\smash{\bm{\xi}_j \sim \mathcal{N}(0, 2\lambda h I)}$.
\item \textbf{Proximal substep.} The second substep is the push-forward of $\smash{\nu^{k+\frac{1}{2}}}$ by $\smash{\mathrm{prox}_{hJ}(\vx) =}$  $\smash{\argmin_{\vy} J(\vy) + \tfrac{1}{2h}\|\vy - \vx\|^2}$; on a population it maps each member to its proximal point.
\end{enumerate}
Under suitable regularity assumptions, the composition of the two is a first-order consistent discretization of the limit dynamics of (\ref{eq: jko}), and its iterates converge to it as $h \to 0$ (Appendix~\ref{app: step}).
\end{proposition}

Two key consequences follow. 
First, substep (i) re-expands the population with variance $2\lambda h$.
Second, substep (ii) requires evaluating $J$ at the proximal points, potentially necessitating additional rollouts.
To avoid these, we restrict the proximal step to candidates with known costs.
Let $\smash{\gB = \{\vb_i\}_{i=1}^M \supseteq \{\vx_j\}_{j=1}^P}$ denote these candidates, with costs $\smash{J_i := J(\hat{\vs}_0, \vb_i)}$, indexed such that $\smash{\vb_j = \vx_j}$ for $j \le P$.
\revisionarxiv{Moreover, its inclusion allows members with no preferable alternatives to remain in place, ensuring that the step cannot increase the population cost (Lemma~\ref{lem: descent}).}
Restricted to $\gB$, substep (ii) reduces to an assignment problem.

\begin{proposition}[The proximal substep on evaluated candidates]\label{prop: assignment}
Let $\vx^0$ be the perturbed population from substep (i) of Proposition~\ref{prop: splitting}. When restricted to uniform measures supported on $P$ distinct candidates in $\gB$, substep (ii) is solved by $\vx^+_j = \vb_{\pi(j)}$ with
\vspace{-0.2cm}
\begin{equation}
\pi \in \argmin_{\pi}\ \sum_{j=1}^P \Big[ J_{\pi(j)} + \frac{1}{2h} \big\| \vx^0_j - \vb_{\pi(j)} \big\|^2 \Big],
\label{eq: assignment}
\vspace{-0.2cm}
\end{equation}
over injections $\pi : \{1, \dots, P\} \hookrightarrow \{1, \dots, M\}$, a rectangular linear assignment. 
For large $h$, (\ref{eq: assignment}) selects the $P$ cheapest candidates and pairs them to the members by the $W_2$-optimal permutation.
\end{proposition}

\revisionarxiv{The restriction to $\gB$ costs an error bounded in Appendix~\ref{app: restriction} (Lemma~\ref{lem: restriction}), and the step still descends: the targets are no more expensive than the population, on average up to the price of the perturbation and, for $\bm{\xi} = 0$, in every order statistic of the costs (Lemma~\ref{lem: descent}).}

\vspace{-0.2cm}
\paragraph{Why introduce learning?}
At deployment, only the current population has known costs, so $\gB = \gA^k$ and (\ref{eq: assignment}) can only rearrange its members (reducing to the identity when $\bm{\xi} = 0$). Thus, the restricted proximal step cannot improve the population without additional rollouts. During training, a larger rollout budget provides additional evaluated candidates $\gB_T \supset \gA^k$, enabling (\ref{eq: assignment}) to produce improved targets (Lemma~\ref{lem: descent}). We will therefore train $\Phi_\theta$ to map the population $\vx^0$ to these targets $\vx^+$ using only the population and its costs, the information available at deployment.

\subsection{Optimizing Action Populations via Conditional Flows}
\label{sec:method-flow}
\vspace{-0.1cm}
Motivated by the transport formulation of Section \ref{sec: method-transport}, we model population updates as continuous flows in action space, providing a flexible parameterization of how candidates move and redistribute.
Specifically, we represent $\Phi_\theta$ as the time-one map of a conditional velocity field $v_\theta$:
\begin{equation}
\dot{\vx}(\tau) = v_\theta\big(\vx(\tau), \tau;\, \gC^k\big), \qquad \vx(0) = \gA^k + \bm{\xi}^k, \qquad \gA^{k+1} = \vx(1),
\label{eq: update-flow}
\end{equation}
over flow time $\smash{\tau \in [0,1]}$, where $\smash{\xi_j^k \sim \mathcal{N}(0,\sigma_\xi^2 I)}$ with $\smash{\sigma_\xi^2 = 2\lambda h}$.
The perturbation corresponds to the heat substep of Proposition~\ref{prop: splitting}, while the learned flow approximates the proximal substep.
Throughout integration, the evaluated population $\gA^k$ and its scores $\gC^k$ form a fixed scored memory, queried by the transported members at their current positions without additional rollouts.

During deployment, starting from an initial population $\gA^0$, the planner alternates evaluation and transport for $K$ iterations, each transport integrating (\ref{eq: update-flow}) with $N_\tau$ explicit Euler steps,
\begin{equation}
\vx^{(m+1)} = \vx^{(m)} + 1 / N_\tau \cdot v_\theta\big(\vx^{(m)}, \tfrac{m}{N_\tau};\, \gC^k\big), \qquad m = 0, \dots, N_\tau - 1,
\label{eq:euler}
\end{equation}
with $\vx^{(0)} = \vx^k + \bm{\xi}^k$ and $\vx^{k+1} = \vx^{(N_\tau)}$. Before $\vx^{k+1}$ is evaluated, the cheapest candidate found so far replaces the transported copy of the highest-cost member. After $K$ iterations the planner returns $\bar{\va}^\star$ of (\ref{eq:planner_output}), at a total cost of $(K+1)P$ rollouts. In receding-horizon control, the first action of $\bar{\va}^\star$ is applied and the next control step is warm-started by shifting $\gA^K$ one step forward.

Several key properties follow.
First, the context remains fixed during integration, so each iteration requires only $P$ rollouts, regardless of $N_\tau$.
Second, rather than compressing the population into a parametric distribution, the velocity field transports the members.
Finally, the field is conditioned on rollout costs via $\gC^k$ rather than the current state $s^k$, allowing the planner to adapt to the deployment objective.
Thus, FlowOpt Policy learns \emph{how to optimize} rather than \emph{what to output}.

\vspace{-0.05cm}
\subsection{Training}
\label{sec: training}
\vspace{-0.1cm}
Training the velocity field (\ref{eq: update-flow}) faces two challenges.
First, we assume no demonstration data and use rollout cost as a signal of candidate quality.
Second, the populations encountered in closed loop are induced by the method itself.
To address both, we run the learned optimizer as a receding-horizon controller and train the field on populations labeled by a strong sampling-based teacher.

The initial population of each control step is the final population of the previous step shifted, while the first step of an episode draws $\gA^0$ with $\bar{\va}_j \sim \mathcal{N}(\mathbf{0}, \sigma_{\mathrm{init}}^2 \mI)$. The same procedure is used in training and deployment, so training populations resemble those encountered at deployment.
Each population $\vx^k$ met by the field at state $\hat{\vs}_n$ is labeled by a strong sampling-based optimizer that is initialized at $\vx^k$ run for $K_T$ iterations of $P_T$ rollouts. 
The teacher's evaluated set $\gB^k_T$ consists of $\vx^k$ and in addition, the candidates evaluated by the teacher. 
The training source is $\smash{\vx^{k,0}=\vx^k+\bm{\xi}^k}$ from (\ref{eq: update-flow}), and the target $\smash{\vx^{k,+}}$ is given by the assignment (\ref{eq: assignment}) on $\gB_T^k$.
By Lemma~\ref{lem: descent}, the mean target cost exceeds the original by at most the perturbation penalty.
For large $h$, (\ref{eq: assignment}) selects the $P$ cheapest candidates and matches them to the sources by $W_2$-optimal transport. 
Each tuple $(\vx^k, \gC^k, \vx^{k,0},\, \vx^{k,+})$ is stored, together with $\gB^k_T$ and its costs.
Every $N_E$ episodes the field is refitted on all stored tuples with the flow-matching loss on straight paths $\vx_\tau = (1-\tau)\vx^0 + \tau \vx^+$,
\begin{equation}
\mathcal{L}(\theta) = \mathbb{E}_{(\vx, \gC, \vx^0, \vx^{+}) \sim \mathcal{D},\; \tau \sim U[0,1)}\; \frac{1}{P} \sum_{j=1}^P \Big\| v_\theta\big(\vx_\tau, \tau;\, \gC\big)_j - \big(\vx^{+}_j - \vx^0_j\big) \Big\|^2,
\label{eq: fm loss}
\end{equation}
which requires no simulator calls, since $\gC$ is rebuilt from stored rollouts. The field thus learns to mimic the performance of $K_T$ teacher iterations at the cost of a single population evaluation. 
By Proposition~\ref{prop: what the field learns}, the optimal velocity is the conditional mean source-to-target displacement given the interpolated population, flow time and context. Under member-wise conditioning, this takes the posterior-weighted form of Theorem~\ref{thm: attention}(i) given the teacher's evaluated set. Moreover, the assignment yields $W_2$-optimal pairings and geodesic, non-crossing paths (Proposition~\ref{prop: straight paths}, Appendix~\ref{app: training}).

\vspace{-0.1cm}
\subsection{Architecture of the Velocity Field}
\label{sec: architecture}
\vspace{-0.1cm}
The natural next question is \textit{what neural network architecture} is the most suitable. There is three key desiderata here. First, the learned optimizer should be invariant to the scale and shifts of the action sequences and costs, since the velocity field in (\ref{eq: update-flow}) operates on populations ranging from a random cloud at the start of an optimization to a tight cluster near a solution. Second, each candidate should observe the other candidates together with their costs and derive its update from the most promising combination of them. Third, the architecture should be permutation equivariant among the members. 
\vspace{-0.5cm}
\paragraph{Transformation invariance.}
The scores assigned to the evaluated candidates is the key signal that directs the transport, yet raw costs might be a poor conditioning signal for the velocity field, as their scale varies across instances and iterations.
We therefore describe each member relative to its population, through a normalized rank $\smash{\rho^k_j = (\mathrm{rank}(J^k_j)-1)/(P-1)}$ and a standardized cost $\smash{\tilde J^k_j = (J^k_j - \bar J^k)/\sigma^k}$, with $\smash{\bar J^k}$ and $\sigma^k$ the empirical mean and standard deviation. As a result, the iteration context $\smash{\gC^k = \{(\rho^k_j, \tilde J^k_j)\}_{j=1}^P}$ is invariant to positive affine rescalings $J \mapsto aJ + b$, $a > 0$. 
Similarly, actions are handled relative to the population:
with $\bar{\vx}^k$ and $s^k$ denoting its mean and
root-mean-square deviation, the network reads members
in standardized coordinates and returns its output
in the same units,
$\smash{v_\theta \big(\vx(\tau), \tau;\, \vx^k, \gC^k ) = s^k \, g_\theta (\hat{\vx}(\tau), \tau;\, \hat{\vx}^k, \gC^k
\big)}$, with
$\smash{\hat{\vx}_j = (\vx_j - \bar{\vx}^k)/s^k}$.
The temperature and step size are scaled accordingly,
$\smash{\lambda = \tilde{\lambda}} \sigma^k$,
$\smash{h = \tilde{h}\, (s^k_\xi)^2 / \sigma^k}$,
$\smash{\sigma_\xi^2 = 2 \tilde{\lambda} \tilde{h}\, (s^k_\xi)^2},
$
with $\smash{s_\xi^k=\operatorname{clip}(s^k,s_{\min},\sigma_{\mathrm{init}})}$,
where $s_{\min}\propto\sigma_{\mathrm{init}}$, bounds the population spread,
so that $\smash{(\tilde{\lambda},\tilde{h})}$ are dimensionless 
and the perturbation in (\ref{eq: update-flow}) shrinks with the population.
The step is invariant to positive affine cost rescaling and equivariant to similarities of the action space applied jointly to $\sigma_{\mathrm{init}}$ (Proposition~\ref{prop: cov}).

\vspace{-0.2cm}
\paragraph{Attention over the population.}
The evaluated population acts as a scored memory from which transported members gather information.
A shared embedding pairs each action sequence with its scores, i.e., $\smash{\vh_i^0 = \eta_\theta(\hat{\vx}_i^k, \rho_i^k, \tilde{J}_i^k)}$ for each evaluated member and $\smash{\vh_j = \eta_\theta(\hat{\vx}_j(\tau), \rho_j^k, \tilde{J}_j^k)}$ for each transported member at its current position.
Then, a cross-attention layer allows each transported member to attend to the entire evaluated population, while a shared decoder $\psi_\theta$ returns its normalized velocity,
$\smash{g_\theta(\cdot)_j = \psi_\theta\big( \vh_j + \operatorname{Attn}_\theta\big(\vh_j;\, \{\vh_i^0\}_{i=1}^P\big),\; \tau
\big)
}$.
Pairing $\hat{\vx}_i^k$ with $(\rho_i^k, \tilde{J}_i^k)$ inside $\vh_i^0$ associates each score with its evaluated candidate, while the residual preserves each member's own position and score alongside the attended information. 
Thus, each member's velocity depends on both its own state and the entire scored population, rather than compressing the latter into a single weighted aggregate.
The architecture is permutation equivariant, since attention is invariant to the ordering of the evaluated members, the embedding and decoder share weights across members, and $\smash{(\bar{\vx}^k, s^k)}$ are symmetric population statistics.
The following theorem shows that attention is not an arbitrary architectural choice, but arises naturally from our flow matching and proximal formulations. First, the optimal flow-matching field for targets supported on evaluated candidates involves a softmax-weighted aggregation of these targets. Second, an entropy-regularized relaxation of the proximal step admits an exact representation through dot-product attention.

\begin{theorem}[Attention as a natural parameterization]
\label{thm: attention}
Let $\smash{\gB=\{\vb_i\}_{i=1}^M}$ be a set of evaluated candidates with costs $J_i$. For each population member $j$, let the source be $\smash{\vx_j^0\sim\mathcal{N}(\vx_j,\sigma_\xi^2 I)}$ and the target $\smash{\vx^+_j} \in \gB$. Let $g_{ji}(\vu)$ be the probability that
(\ref{eq: assignment}) assigns member $j$ to candidate $\vb_i$
given its source $\vx_j^0 = \vu$.
\begin{enumerate}[label=(\roman*), leftmargin=2.2em, itemsep=0pt, topsep=2pt]
\item \textbf{Optimal velocity field.}
For the straight paths $\smash{\vx_\tau = (1-\tau)\vx^0_j + \tau\vx^+_j}$, the optimal velocity field for member $j$, i.e., the minimizer of the flow-matching loss, is given on $\tau \in [0,1)$ by
\vspace{-0.1cm}
\begin{equation}
v^\star(\vz,\tau)
=\frac{1}{1-\tau}
\bigg( \sum_{i=1}^M\alpha_i\vb_i-\vz \bigg),
\quad
\alpha_i(\vz,\tau)=\operatorname{softmax}_i\big(\langle\vq,\vb_i\rangle+\beta_i+\log g_i(\vu_i)\big)
\label{eq: posterior_field}
\end{equation}
where $\vu_i=(\vz-\tau\vb_i)/(1-\tau)$ is the
source position consistent with $\vz$ and target $\vb_i$,
while
$\smash{\vq=\tau[\vz-(1-\tau)\vx_j]/((1-\tau)^2\sigma_\xi^2)}$
and
$\smash{\beta_i=-\tau^2\|\vb_i\|^2/(2(1-\tau)^2\sigma_\xi^2)}$.
Thus, the optimal velocity is a softmax-weighted combination of candidate targets, followed by a residual subtraction and scaling by $1/(1-\tau)$.
When $g_i$ is independent of the source, the weights reduce to dot-product attention with query $\vq$, keys and values $\vb_i$, and additive key biases $\beta_i+\log g_i$.

\item \textbf{Entropic proximal step.}
Let $\nu_0$ be a strictly positive reference distribution
on $\gB$, and let $h,\lambda>0$.
For a source $\vx$, consider the entropic
relaxation of the proximal assignment (\ref{eq: assignment}),
\vspace{-0.1cm}
\begin{equation}
\min_{\alpha\in\Delta_M}
\sum_{i=1}^M\alpha_i
\left[J_i+\frac{\|\vx-\vb_i\|^2}{2h}\right]
+\lambda\KL(\alpha\|\nu_0).
\label{eq: soft step}
\end{equation}
Its unique minimizer is
$\smash{\alpha_i^\star(\vx)=\operatorname{softmax}_i(\langle\vq,\vb_i\rangle+\beta_i)}$,
where $\vq=\vx/(h\lambda)$ and
$\beta_i=\log\nu_{0,i}-J_i/\lambda
-\|\vb_i\|^2/(2h\lambda)$.
Thus, the soft proximal step
$\sum_i\alpha_i^\star(\vx)\vb_i$
is a cross-attention operation with query $\vq$, keys and values $\vb_i$, and additive key biases $\beta_i$.
\end{enumerate}
\end{theorem}

At $\tau=0$, the two characterizations agree when the teacher samples targets from the relaxed step (\ref{eq: soft step}), since $g_i=\alpha_i^\star$ gives $v^\star(\vz,0)=\sum_i\alpha_i^\star(\vz)\vb_i-\vz$. Thus, the initial flow-matching velocity is the displacement prescribed by the soft proximal step.

\revisionarxiv{
Part (i) is the layer $\operatorname{Attn}_\theta$ up to the source-dependent term $\log g_i$, which the learned projections approximate; part (ii) says that learning the projections learns the metric, step size and temperature of the step as functions of the context, with MPPI and CEM as the query-independent corner (Corollary~\ref{cor: corners}).
}

Our attention-based parameterization is closely related to the core mechanism underlying classical sampling-based optimizers such as MPPI and CEM. These methods assign greater influence to promising action sequences through their rollout costs and aggregate across the resulting weighted population, which can be viewed as a simplified operation performed by our member-axis attention. 
{Theorem~\ref{thm: attention}(ii) makes this connection precise: the soft proximal step takes the form of an attention layer whose logits combine a transport term, scaled by the step size, with a cost term. As $h \to \infty$, the transport term vanishes, leaving cost-based weights that are identical across population members.}

\begin{corollary}[MPPI and CEM as special attention patterns]\label{cor: corners}
Let the values be the candidate action sequences, $\vv_i = \bar{\va}_i$. With query-independent logits $r_{ji} = \beta_i$ and a second head returning the plain mean, one Euler step of the attention block implements $\smash{\bar{\va}_j^+ = \bar{\va}_j + \big(\sum_{i=1}^P w_i \bar{\va}_i - \frac{1}{P}\sum_{i=1}^P \bar{\va}_i\big)}$ with $w = \operatorname{softmax}(\beta)$ for every member $j$, so the mean of the updated population equals the weighted mean $\sum_i w_i \bar{\va}_i$. Setting $\beta_i = -J_i/\lambda$ yields the MPPI weights, and $\beta_i = 0$ for $i \in \gE$ and $-\beta$ otherwise yields the CEM weights $\mathbf{1}\{i \in \gE\}/|\gE|$ as $\beta \to \infty$. {This corresponds to the limit $h \to \infty$ in Theorem~\ref{thm: attention}(ii), where the transport terms vanish and the weights depend only on the costs. Population re-expansion is then handled separately, e.g., through MPPI's covariance $\bm\Sigma$.}
\end{corollary}

The above exposition reveals that MPPI and CEM correspond to restricted attention patterns with cost-based logits identical across members, whereas our learned attention relaxes both restrictions. The logits can depend jointly on the member being updated and the members supplying information, using their actions and rollout costs. 
By Theorem~\ref{thm: attention}(ii), learning the projections amounts to learning the metric, step size and temperature as functions of the context.
Consequently, different members may draw information from different promising regions of the population, rather than all being driven toward a single average. Our parameterization therefore retains the clasical cost-weighted aggregation, while enabling richer, context-dependent population updates.

\vspace{-0.1cm}
\section{Experiments}
\vspace{-0.2cm}
Our experiments span from classical physics-based systems to learned latent world models. The results highlight that \flowoptpolicy significantly outperforms handcrafted and learned sampling-based planners at an equal budget, and that optimizing at deployment yields better performance and generalization under distribution shift than widely used direct state-conditioned action generation.

\newcommand{\ms}[2]{#1{\scriptsize$\,\pm\,$#2}}
\newcommand{\na}{--}

\begin{figure}[t]
\vspace{-0.2cm}
  \begin{center}
    \centerline{\includegraphics[width=\textwidth]{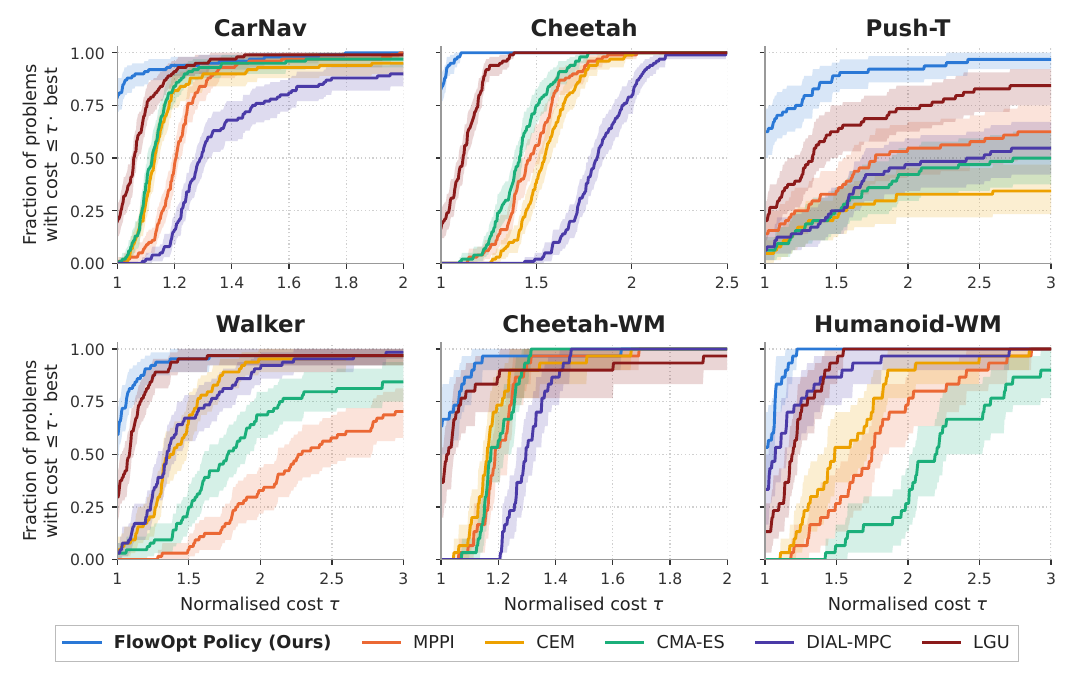}}
    \vspace{-0.5cm}
\caption{\textbf{Performance profiles on physics-based and world-model tasks.} Each curve shows the fraction of problems on which a method's closed-loop cost is within a factor $\tau$ of the best performer (closer to the top-left is better). \flowoptpolicy~is at or near the best cost on nearly all problems.}
    \label{fig: performance profiles}
  \end{center}
\vspace{-0.8cm}
\end{figure}

\vspace{-0.1cm}
\subsection{Tasks and Baselines}
\vspace{-0.15cm}

\textbf{Tasks and metrics.} We consider four physics-based tasks, CarNav (navigation among obstacles with moving targets), HalfCheetah and Walker from MuJoCo \citep{todorov2012mujoco}, and Push-T \citep{chi2025diffusionpolicy}, as well as HalfCheetah-WM and Humanoid-WM planned inside the TD-MPC2 world model \citep{hansen2024tdmpc2}. On the physics tasks, all planners roll out the ground-truth simulator dynamics, while on the world-model tasks, they share the same frozen latent dynamics and cost, evaluated from the current encoded observation.
All methods are evaluated in closed loop on held-out instances shared across methods. We report cost for CarNav and Walker, return for HalfCheetah and the world-model tasks, and success rate for Push-T, with costs and returns normalized by the best method on each task. Full metric definitions and details on the construction of the performance profiles are given in Appendix~\ref{sec: appendix experiment details}.

\textbf{Baselines.} We compare \flowoptpolicy against well-established sampling-based methods, namely \mppi~\citep{williams2017information}, \cem~\citep{deboer2005cem_tutorial}, \cmaes~\citep{hansen2001cmaes}, and \dialmpc~\citep{xue2025dial}. 
We additionally consider two learned baselines.
To assess the benefits of transporting the population directly rather than updating a Gaussian sampling distribution, we introduce a \lgufull~baseline inspired by \citet{sacks2022learning}, which learns such an update from evaluated candidates and their costs.
We also consider \flowpolicy, a generative policy conditioned on the current state and available task information \citep{chi2025diffusionpolicy, black2024pi_0}, which reveals the value of optimizing through a predictive model instead of generating actions directly.
Both baselines use the same training procedure and network backbone as \flowoptpolicy.
All model-based planners use the same number of rollouts (Appendix~\ref{sec: appendix experiment details}).

\vspace{-0.6cm}
\subsection{Comparison with Sampling-Based Planners}
\label{sec: experiments - comp with sampling based}
\vspace{-0.15cm}
Figure~\ref{fig: performance profiles} compares \flowoptpolicy with the classical planners and \lgu~through performance profiles \citep{dolan2002benchmarking}, which show how consistently each method performs across individual episodes, while aggregate metrics are given in Table~\ref{tab: main - metrics} in Appendix~\ref{sec: appendix experiment details}.
\flowoptpolicy achieves the highest performance profiles across all six tasks, consistently reaching near-optimal costs at smaller $\tau$ than the baselines. \lgu~is overall the strongest baseline, with the largest margins over the non-learned methods on HalfCheetah, Walker, Push-T and Humanoid-WM. The advantages of \flowoptpolicy are especially pronounced on Push-T, where it achieves a $99.0\%$ success rate compared with $86.5\%$ for \lgu, with its performance profile remaining considerably higher.
Since all planners roll out the same model under the same budget, the gains over the classical planners stem from learning the update, while the gains over \lgu, which shares our backbone and training, stem from transporting candidates directly rather than updating a Gaussian sampling distribution.

\vspace{-0.1cm}
\subsection{Generalization to Unseen Objectives}
\label{sec: experiments - generalization}
\vspace{-0.15cm}

\begin{figure}[t]
  \begin{center}
    \centerline{\includegraphics[width=\textwidth]{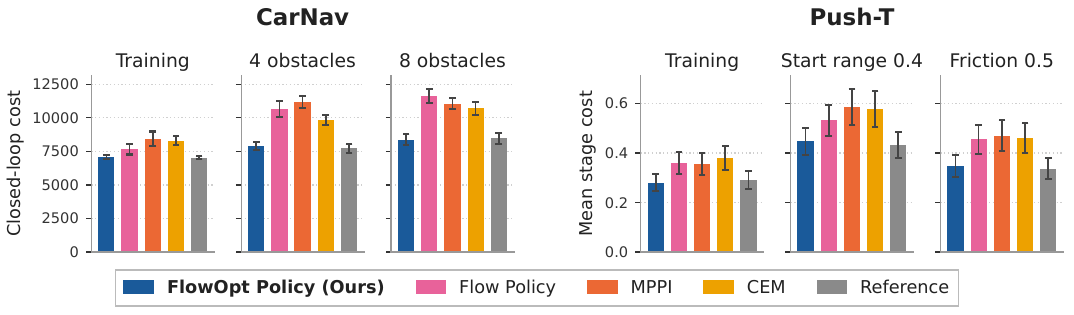}}
    \vspace{-0.4cm}
\caption{\textbf{Generalization capabilities.} \flowoptpolicy against \flowpolicy, \mppi, \cem, and a higher-budget reference optimizer. We evaluate CarNav with additional obstacles and Push-T with a wider initial block range and altered friction. \flowoptpolicy consistently outperforms the other methods under distribution shift, approaching the performance of the higher-budget reference.}
    \label{fig: vs flow policy}
  \end{center}
\vspace{-0.6cm}
\end{figure}

Next, we compare against \flowpolicy, which shares the architecture and training with \flowoptpolicy but directly outputs state-conditioned action sequences.
We also include equal-budget \mppi~and \cem, and the higher-budget teacher as reference.
We evaluate all methods on the training distribution and on shifted settings (Figure~\ref{fig: vs flow policy}). On CarNav, \flowoptpolicy slightly outperforms \flowpolicy~in distribution and achieves much better costs with four and eight obstacles instead of three. Similarly, on Push-T, \flowoptpolicy is better in distribution as well as with a wider initial block range and altered friction. Across both tasks, our method also outperforms \mppi~and \cem~and approaches the higher-budget reference. 
When a control-rate penalty absent from training is added to Push-T (Table~\ref{tab:objective_shift}), \flowoptpolicy adapts its actions without retraining, whereas \flowpolicy~cannot respond to the new objective.
\flowoptpolicy generalizes better since the shifts are reflected in the rollout costs, which are standardized before entering the velocity field, whereas for \flowpolicy, they directly affect the state inputs, which hurts generalization.

\begin{table}[t]
\centering
\vspace{-0.3cm}
\caption{Component ablations on CarNav and Push-T, with differences relative to the full model.}
\label{tab:component_ablations}
\vspace{0.15cm}
\footnotesize
\setlength{\tabcolsep}{4pt}
\begin{tabular*}{\linewidth}{@{}l@{\extracolsep{\fill}}lrrrr@{}}
\toprule
& & \multicolumn{2}{c}{\textbf{CarNav}}
    & \multicolumn{2}{c}{\textbf{Push-T}} \\
\cmidrule(lr){3-4}\cmidrule(lr){5-6}
\textbf{Ablation} & \textbf{Variant}
& \textbf{Cost} & \textbf{$\Delta$ vs. full}
& \textbf{SR (\%)} & \textbf{$\Delta$ vs. full} \\
\midrule
-- & Full model & 7095 & -- & 99.0 & -- \\
\midrule
Population attention & No attention
    & 7512 & $+418 \pm 110$ & 68.8 & $-30.2 \pm 4.8$ \\
Cost ranking & Z-score only
    & 7116 & $+21 \pm 115$ & 83.9 & $-15.1 \pm 3.3$ \\
Cost standardization & Rank only
    & 7169 & $+75 \pm 102$ & 92.7 & $-6.3 \pm 4.1$ \\
\midrule
Flow integration & 1 Euler step
    & 8079 & $+984 \pm 168$ & 49.0 & $-50.0 \pm 6.2$ \\
Flow integration & 2 Euler steps
    & 7583 & $+488 \pm 109$ & 71.9 & $-27.1 \pm 5.5$ \\
Flow integration & 10 Euler steps
    & 6966 & $-129 \pm 77$ & 98.4 & $-0.5 \pm 0.9$ \\

\midrule
Teacher quality & 4$\times$ fewer rollouts
    & 7193 & $+98 \pm 86$ & 94.3 & $-4.7 \pm 1.6$ \\
Teacher quality & 16$\times$ fewer rollouts
    & 7278 & $+183 \pm 91$ & 89.6 & $-9.4 \pm 3.1$ \\
\midrule
Network size & Width 64 / 256
    & 7378 & $+283 \pm 103$ & 93.8 & $-5.2 \pm 1.8$ \\
Network size & Width 384 / 1024
    & 6833 & $-261 \pm 123$ & 99.0 & $+0.0 \pm 1.6$ \\
\bottomrule
\end{tabular*}
\vspace{-0.4cm}
\end{table}

\subsection{Ablations}\label{sec: experiments - ablations}

We investigate the contributions of individual components of \flowoptpolicy on CarNav and Push-T (Table~\ref{tab:component_ablations}).
Removing population attention substantially increases cost, highlighting the importance of the candidates attending to each other.
Flow integration has the largest effect, as fewer Euler steps degrade performance, while ten steps offer only a marginal improvement over the default five.
Using only ranks or standardized costs matters little on CarNav but lowers the Push-T success rate by up to $15$ points.
Performance degrades gracefully with weaker teachers and improves with a wider network.
We also examine the effect of increasing the number of optimization iterations beyond the training budget (Figure~\ref{fig:ablation} in Appendix~\ref{sec: appendix experiment details}). On both CarNav and Push-T, \flowoptpolicy with three iterations outperforms \mppi~and \cem~with nine. Additional iterations further improve performance on CarNav without degrading it on Push-T, while the baselines improve but do not close the gap.

\paragraph{Limitations.} 

\flowoptpolicy is essentially a learning-to-optimize method so although it is shown to generalize better than direct generative policies, its learned transport still remains primarily tailored to the training distribution. The current training scheme is potentially myopic, since the teacher supervises a single population update over a finite horizon rather than the closed-loop return; therefore combining the optimizer with a learned terminal value function is a natural next direction. 
Finally, as the method drives populations toward low-cost regions more aggressively, it may exploit inaccuracies of a learned world model in regions where the model was never validated.
\vspace{-0.2cm}

\section{Conclusion}
We introduced FlowOpt Policy, which casts planning as a learned transport problem over action sequences and learns, via a flow-based model parameterized by attention over the population, how candidates should move given how they scored.
We further show that classical sampling-based planners are recovered as special cases through fixed attention patterns.
Across physics-based and world-model tasks, FlowOpt Policy consistently outperforms sampling-based planners, and generalizes better than state-conditioned generative policies to unseen objectives.
We believe that this work paves the way toward planners that learn their own optimization procedure, combining the adaptability of model-based search with the expressiveness of modern generative models.

\newpage
\bibliography{iclr2027_conference}

\newpage
\appendix

{
\renewcommand{\contentsname}{Table of Contents}
\setcounter{tocdepth}{2} %
\tableofcontents
}

\newpage
\section{Preliminaries}

\subsection{Lifting to measures and free-energy formulation}\label{app: theory - lift}

This section supports the paragraph ``Lifting to measures and free-energy formulation'' of Section~\ref{sec: method-transport}. 
Lemma~\ref{lem: lift} recalls and proves for completeness three classical claims: 
first, that minimizing the expected cost over measures preserves the optimal value of (\ref{eq: opt_problem}), second, that the minimizer of (\ref{eq: free energy}) is the Gibbs measure $\nu_\lambda$, and last, that $\nu_\lambda$ concentrates on the minimizers of $J$ as $\lambda \to 0$.
Since the dynamics clip actions to the box $[\va_{\min}, \va_{\max}]^H$, $J$ is bounded, and we pose (\ref{eq: free energy}) on the box, so that $Z_\lambda < \infty$ and the sublevel sets of $J$ are compact.

\begin{lemma}[Lifting to measures and properties of the free energy]\label{lem: lift}

Let $J : \R^d \to \R$ be measurable and bounded below, and let $J^\star := \inf_{\bar{\va}} J(\hat{\vs}_0, \bar{\va})$.
\begin{enumerate}[label=(\roman*), leftmargin=2.2em, itemsep=0pt, topsep=2pt]
\item \textbf{Lifting.} $\inf_{\nu \in \mathcal{P}(\R^d)} \E_\nu[J] = J^\star$. The infimum is attained if and only if $J^\star$ is, and then the minimizers are exactly the measures supported on $\argmin J$.
\item \textbf{Gibbs minimizer.} If $\smash{Z_\lambda := \int e^{-J/\lambda} \rd\bar{\va} < \infty}$, then $F_\lambda(\nu) = \lambda \KL(\nu \,\|\, \nu_\lambda) - \lambda \log Z_\lambda$ for every $\nu$, with $\smash{\nu_\lambda := e^{-J/\lambda}/Z_\lambda}$. Hence $\nu_\lambda$ is the unique minimizer of (\ref{eq: free energy}) and $\smash{\min_\nu F_\lambda = -\lambda \log Z_\lambda}$.
\item \textbf{Zero-temperature limit.} If moreover $J$ is continuous with compact sublevel sets, then $-\lambda \log Z_\lambda \to J^\star$ and $\nu_\lambda$ converges weakly, as $\lambda \to 0$, to a probability measure supported on $\argmin J$.
\end{enumerate}
\end{lemma}

\begin{proof}
(i) Since $J \ge J^\star$ pointwise, it follows that $\E_\nu[J] \ge J^\star$ for every $\nu$. 
Conversely, for $\epsilon > 0$, pick $\bar{\va}_\epsilon$ with $J(\bar{\va}_\epsilon) \le J^\star + \epsilon$, so that $\E_{\delta_{\bar{\va}_\epsilon}}[J] \le J^\star + \epsilon$. If $\E_\nu[J] = J^\star$ then $J = J^\star$ $\nu$-almost everywhere, i.e.\ $\nu(\argmin J) = 1$, and conversely. 

(ii) For $\nu$ with density $\rho$,
\begin{equation}
\lambda \KL(\nu \,\|\, \nu_\lambda) = \lambda \int \rho \log \rho \,\rd\bar{\va} + \int J \rho \,\rd\bar{\va} + \lambda \log Z_\lambda = F_\lambda(\nu) + \lambda \log Z_\lambda ;
\end{equation}
for $\nu$ without a density both sides equal $+\infty$. Since $\KL \ge 0$ with equality if and only if $\nu = \nu_\lambda$, the claim follows (the Gibbs variational principle, \citealp[Prop.~1.4.2]{dupuis2011weak}).

(iii) The first statement is Laplace's principle, $\smash{\lambda \log \int e^{-J/\lambda} \rd\bar{\va} \to -J^\star}$; the second is \citet{hwang1980laplace}.
\end{proof}

\paragraph{Priors.} For a prior density $p$ on the box, $\smash{\lambda D_{\mathrm{KL}}(\nu \,\|\, p) = \lambda \mathcal{H}(\nu) - \lambda \E_\nu[\log p]}$, so the free energy of MPPI \citep{williams2017information}, which takes the entropy relative to the Gaussian proposal $p$ around the nominal sequence, is (\ref{eq: free energy}) with $J$ replaced by $J - \lambda \log p$, and its minimizer $\smash{\propto p\, e^{-J/\lambda}}$ is biased toward $p$. We take the Lebesgue reference because the target of the planner is problem (\ref{eq: opt_problem}) itself. A learned policy prior, as in TD-MPC2 \citep{hansen2024tdmpc2}, enters the same way, and Appendices~\ref{app: jko step} and~\ref{app: step} go through unchanged with $J - \lambda \log p$ in place of $J$.

\subsection{JKO scheme}\label{app: jko step}

Throughout this section, $J$ is lower semicontinuous and bounded below, and all measures are defined on $\R^d$. On the action box the same arguments hold once the heat flow is confined to the box, that is, with no flux of mass through its boundary.

For general measures $\nu, \nu' \in \mathcal{P}_2(\R^d)$, the Wasserstein distance is defined through couplings,
\begin{equation}
W_2^2(\nu, \nu') := \min_{\gamma \in \Gamma(\nu, \nu')} \int \|\vx - \vy\|^2 \, \rd\gamma(\vx, \vy),
\label{eq: w2}
\end{equation}
where $\Gamma(\nu, \nu')$ is the set of probability measures on $\R^d \times \R^d$ with marginals $\nu$ and $\nu'$. 
For two populations of the same size, the minimum in (\ref{eq: w2}) is attained at a permutation, since the couplings are then the doubly stochastic matrices scaled by $1/P$, whose extreme points are the permutation matrices (Birkhoff--von Neumann); this gives the permutation formula of Section~\ref{sec: method-transport}.

\begin{lemma}[Existence, regularity and descent of the JKO step]\label{lem: jko}
Given $\nu^k \in \mathcal{P}_2(\R^d)$, the JKO step (\ref{eq: jko}) has a unique minimizer $\nu^{k+1}$, which is absolutely continuous and satisfies
\begin{equation}
F_\lambda(\nu^{k+1}) + \frac{1}{2h}\, W_2^2(\nu^{k+1}, \nu^k) \le F_\lambda(\nu^k),
\label{eq: jko_descent}
\end{equation}
so the step decreases $F_\lambda$ by $\Delta F_\lambda = F_\lambda(\nu^k) - F_\lambda(\nu^{k+1}) \ge 0$, displacing mass by at most $\sqrt{2h\,\Delta F_\lambda}$.
\end{lemma}

\begin{proof}
We show in turn that (i) the infimum is finite, (ii) a minimizer exists, (iii) the minimizer is unique and has a density, and (iv) it satisfies (\ref{eq: jko_descent}):
\begin{enumerate}
    \item \emph{Finiteness.} The infimum in (\ref{eq: jko}) is trivially less than $+\infty$, since any Gaussian has finite $F_\lambda$ and finite distance to $\nu^k$. The objective is also bounded below, as in \citep[Prop.~4.1]{jordan1998variational}, so the infimum is finite.
    \item \emph{Existence.} Let $\nu_n$ be a sequence whose objective values decrease to the infimum. The term $\frac{1}{2h}W_2^2(\nu_n, \nu^k)$ stays bounded, so the second moments are bounded and the sequence is tight; by Prokhorov's theorem a subsequence converges weakly to some $\bar\nu$. Each term of the objective is lower semicontinuous under weak convergence, namely $\E_\nu[J]$ because $J$ is lower semicontinuous and bounded below, $\mathcal{H}$ on sets of bounded second moment, and $W_2^2(\cdot, \nu^k)$ in general, so the objective at $\bar\nu$ is at most the infimum. Hence $\bar\nu$ is a minimizer.
    \item \emph{Uniqueness and regularity.} The minimizer is unique because $\nu \mapsto W_2^2(\nu, \nu^k)$ and $\nu \mapsto \E_\nu[J]$ are convex along linear interpolations of measures and $\mathcal{H}$ is strictly convex. Its objective value is finite, so $\mathcal{H}(\nu^{k+1}) < \infty$, which means that $\nu^{k+1}$ has a density.
    \item \emph{Descent.} Comparing the minimizer with the admissible choice $\nu = \nu^k$ gives (\ref{eq: jko_descent}).
\end{enumerate}
\end{proof}

\section{Theory and Proofs}
\label{sec: app - proofs}

\subsection{Heat--proximal splitting}\label{app: step}

\begin{proof}[Proof of Proposition~\ref{prop: splitting}]
We establish the two characterizations in turn and then show that their composition approximates the Wasserstein gradient flow of the free energy.

\emph{(i) Heat substep.}
The Wasserstein gradient flow of $\lambda\mathcal{H}$ is the heat equation $\partial_t\rho=\lambda\Delta\rho$ \citep{ambrosio2005gradient}. Starting from any $\nu\in\mathcal{P}_2(\R^d)$, its solution after time $h$ is
\begin{equation}
\nu^{k+\frac12}=\nu*\mathcal{N}(0,2\lambda hI).
\end{equation}
For an empirical measure $\mu_{\vx}=\frac1P\sum_j\delta_{\vx_j}$, this distribution can be sampled by independently perturbing each member, $\vx_j^0=\vx_j+\bm{\xi}_j$, with $\bm{\xi}_j\sim\mathcal{N}(0,2\lambda hI)$. Indeed,
\begin{equation}
\E_{\bm{\xi}}[\mu_{\vx^0}]
=\frac1P\sum_j\mathcal{N}(\vx_j,2\lambda hI)
=\mu_{\vx}*\mathcal{N}(0,2\lambda hI).
\end{equation}
Thus, Gaussian perturbation provides an unbiased empirical representation of the heat substep.

\emph{(ii) Proximal substep.}
Let $\nu^0:=\nu^{k+\frac12}$ and define
\begin{equation}
M_h(\vx):=\min_{\vy}\left\{
J(\vy)+\frac1{2h}\|\vy-\vx\|^2
\right\}.
\end{equation}
Since $J$ is lower semicontinuous and bounded below, the minimum is attained, and $\mathrm{prox}_{hJ}(\vx)$ admits a measurable selection. By the definition of $W_2$ in (\ref{eq: w2}), for any measure $\nu$,
\begin{equation}
\E_\nu[J]+\frac1{2h}W_2^2(\nu,\nu^0)
=\min_{\gamma\in\Gamma(\nu^0,\nu)}
\int\!\left[J(\vy)+\frac1{2h}\|\vy-\vx\|^2\right]\!
\rd\gamma(\vx,\vy)\\
\geq \int\! M_h(\vx)\,\rd\nu^0(\vx).
\end{equation}
The lower bound is attained by sending each $\vx$ to one of its proximal points $\mathrm{prox}_{hJ}(\vx)$. Indeed, choosing any measurable selection gives the minimizing measure
\begin{equation}
\nu^{k+1}=(\mathrm{prox}_{hJ})_\#\nu^0,
\end{equation}
with equality in the bound above. When the proximal map is single-valued $\nu^0$-almost everywhere, this minimizer is unique. In particular, for an empirical measure $\nu^0=\mu_{\vx^0}$, the step acts independently on each member:
\begin{equation}
\nu^{k+1}
=\frac1P\sum_{j=1}^P
\delta_{\mathrm{prox}_{hJ}(\vx_j^0)}.
\end{equation}

\emph{(iii) Consistency and convergence.}
Suppose $J\in C^2$ with globally Lipschitz gradient, and let $h\|\nabla^2J\|_\infty<1$, so the proximal map is single-valued. Combining the two substeps gives the stochastic update
\begin{equation}
X^{k+1}=\mathrm{prox}_{hJ}
\bigl(X^k+\sqrt{2\lambda h}\,Z^k\bigr),
\qquad Z^k\sim\mathcal{N}(0,I).
\end{equation}
By the optimality condition of the proximal map, this is equivalent to
\begin{equation}
X^{k+1}=X^k-h\nabla J(X^{k+1})
+\sqrt{2\lambda h}\,Z^k,
\end{equation}
which is the implicit Euler--Maruyama discretization of the Langevin dynamics
\begin{equation}
\rd X_t=-\nabla J(X_t)\,\rd t
+\sqrt{2\lambda}\,\rd W_t.
\end{equation}
The law $\rho_t$ of this diffusion satisfies the Fokker--Planck equation
\begin{equation}
\partial_t\rho_t
=\nabla\cdot(\rho_t\nabla J)+\lambda\Delta\rho_t,
\end{equation}
which is the Wasserstein gradient flow of $\mathcal{F}_\lambda$ and the continuous-time limit of the JKO scheme (\ref{eq: jko}) \citep{jordan1998variational}.

Under sufficient smoothness, the implicit scheme has the local weak expansion
\begin{equation}
\E[\varphi(X^{k+1})\mid X^k=\vx]
=\varphi(\vx)
+h\bigl(\lambda\Delta\varphi(\vx)
-\nabla J(\vx)\cdot\nabla\varphi(\vx)\bigr)
+O(h^2)
\end{equation}
for smooth test functions $\varphi$, establishing first-order consistency. Moreover, the globally Lipschitz drift ensures convergence of the implicit Euler--Maruyama scheme as $h\to0$, and hence of the measures generated by the splitting to the same gradient flow.
\end{proof}

\subsection{Restriction to evaluated candidates}\label{app: restriction}

This section supports the restriction of the proximal substep to evaluated candidates in Section~\ref{sec: method-transport}, with the proof of Proposition~\ref{prop: assignment} and the descent property of the restricted step (Lemma~\ref{lem: descent}).

\begin{proof}[Proof of Proposition~\ref{prop: assignment}]
We first show that the restricted proximal substep reduces to the assignment problem (\ref{eq: assignment}), and then establish its behavior for large $h$.

Any uniform measure supported on $P$ distinct candidates in $\gB$ can be written as
\begin{equation}
\nu_S=\frac1P\sum_{i\in S}\delta_{\vb_i},
\qquad S\subseteq\{1,\ldots,M\},\quad |S|=P.
\end{equation}
Its expected cost is $\E_{\nu_S}[J]=\frac1P\sum_{i\in S}J_i$. Since both $\nu_S$ and $\mu_{\vx^0}$ are uniform empirical measures with $P$ members, their Wasserstein distance is attained by an optimal one-to-one matching:
\begin{equation}
W_2^2(\nu_S,\mu_{\vx^0})
=\min_{\sigma:\{1,\ldots,P\}\to S}
\frac1P\sum_{j=1}^P
\|\vx_j^0-\vb_{\sigma(j)}\|^2,
\end{equation}
where $\sigma$ ranges over bijections. Substituting into the proximal objective gives
\begin{equation}
\min_{\substack{S\subseteq\{1,\ldots,M\}\\|S|=P}}
\min_{\sigma:\{1,\ldots,P\}\to S}
\frac1P\sum_{j=1}^P
\left[
J_{\sigma(j)}
+\frac1{2h}\|\vx_j^0-\vb_{\sigma(j)}\|^2
\right].
\end{equation}
Choosing a subset $S$ and a bijection onto it is equivalent to choosing an injection $\pi:\{1,\ldots,P\}\hookrightarrow\{1,\ldots,M\}$. Thus, the restricted substep is exactly (\ref{eq: assignment}), with $\vx_j^+=\vb_{\pi(j)}$.

To establish the large-$h$ behavior, write the assignment objective, omitting the constant factor $1/P$,
\begin{equation}
C(\pi)+\frac1{2h}D(\pi),
\quad
C(\pi):=\sum_j J_{\pi(j)},
\quad
D(\pi):=\sum_j\|\vx_j^0-\vb_{\pi(j)}\|^2.
\end{equation}
As $h\to\infty$, the displacement penalty vanishes, leaving the total cost $C(\pi)$. Since there are finitely many possible assignments, for sufficiently large $h$ every minimizer must also minimize $C(\pi)$, and among assignments with the same minimum cost, it minimizes $D(\pi)$. Minimizing $C(\pi)$ amounts to selecting the $P$ cheapest candidates, while minimizing $D(\pi)$ pairs them to the source members through the $W_2$-optimal matching. If candidate costs are tied, the displacement term also determines which of the tied candidates are selected.
\end{proof}

\begin{lemma}[Descent of the restricted step]\label{lem: descent}
Let $\gB \supseteq \{\vx_j\}_{j=1}^P$, $\vx^0 = \vx + \bm{\xi}$, and $\vx^+$ be the targets given by (\ref{eq: assignment}). Then, the following descent inequality holds:
\begin{equation}
\frac{1}{P}\sum_{j=1}^P J(\vx^+_j) + \frac{1}{2hP}\sum_{j=1}^P \|\vx^+_j - \vx^0_j\|^2 \;\le\; \frac{1}{P}\sum_{j=1}^P J(\vx_j) + \frac{1}{2hP}\sum_{j=1}^P \|\bm{\xi}_j\|^2 .
\label{eq: descent}
\end{equation}
Consequently, the mean target cost exceeds the mean population cost by at most $\smash{\frac{1}{2hP}\sum_j \|\bm\xi_j\|^2}$, and the mean squared displacement $\smash{\frac{1}{P}\sum_j \|\vx^+_j - \vx^0_j\|^2}$ is at most $\smash{2h\,\Delta\bar J + \frac{1}{P}\sum_j\|\bm\xi_j\|^2}$, where $\smash{\Delta\bar J := \frac{1}{P}\sum_j J(\vx_j) - \frac{1}{P}\sum_j J(\vx^+_j)}$. If $\bm{\xi} = 0$, every ordered target cost is no greater than the corresponding ordered population cost. In particular, the best cost does not increase.
\end{lemma}

\begin{proof}
Since $\gB$ contains the original population, the identity assignment $\pi_0(j)=j$ is feasible in (\ref{eq: assignment}). Its objective is
\begin{equation}
\sum_{j=1}^P \left[
J_j + \frac{1}{2h}\|\vx_j^0-\vx_j\|^2
\right]
=
\sum_{j=1}^P J_j
+\frac{1}{2h}\sum_{j=1}^P\|\bm{\xi}_j\|^2.
\end{equation}
The optimal assignment cannot have a larger objective, which, after dividing by $P$, gives (\ref{eq: descent}). The bounds on the mean cost and squared displacement follow by rearranging this inequality.

For the final claim, let $\bm{\xi}=0$, so that $\vx^0=\vx$, and compare the optimal assignment $\pi$ with the identity. Since $\pi$ is injective, the members can be partitioned into cycles (including fixed points), on which $\pi$ permutes their original candidates, and chains that introduce a candidate outside the population and leave an original candidate unused. Cycles only rearrange the original costs, so consider a chain of members $j_1,\ldots,j_r$ satisfying
\begin{equation}
\pi(j_1)=e,\qquad
\pi(j_i)=j_{i-1},\quad i=2,\ldots,r,
\end{equation}
where $\vb_e$ is outside the original population and $\vb_{j_r}$ is unused. Along this chain, the original costs $J_{j_1},\ldots,J_{j_r}$ are replaced by $J_e,J_{j_1},\ldots,J_{j_{r-1}}$.

Assigning these members back to their original candidates is feasible, since $\vb_{j_1},\ldots,\vb_{j_{r-1}}$ are used only within the chain and $\vb_{j_r}$ is unused. This reassignment incurs zero displacement cost, so optimality of $\pi$ gives
\begin{equation}
J_e+\frac{1}{2h}\sum_{i=1}^r
\|\vx_{j_i}-\vb_{\pi(j_i)}\|^2
\leq J_{j_r},
\end{equation}
and hence $J_e\leq J_{j_r}$. Thus, each chain replaces one original cost by a no larger one, while cycles leave the costs unchanged. Consequently, every ordered target cost is no greater than the corresponding ordered population cost.
\end{proof}

\begin{revisionarxivblock}
\begin{lemma}[Error of the restriction]\label{lem: restriction}
Let $J$ be $\kappa$-weakly convex, i.e., $J + \frac{\kappa}{2}\|\cdot\|^2$ convex, and let $h < 1/\kappa$, so that $\vy^\star_j := \mathrm{prox}_{hJ}(\vx^0_j)$ is unique. Let $f_j(\vy) := J(\vy) + \frac{1}{2h}\|\vy - \vx^0_j\|^2$, let $\vx^+$ be the targets (\ref{eq: assignment}), and let $\epsilon_{\gB} := \frac{1}{P}\sum_j [ f_j(\vx^+_j) - f_j(\vy^\star_j) ] \ge 0$ be the gap between the restricted and the exact substep objectives. Then
\begin{equation}
\frac{1}{P}\sum_{j=1}^P \|\vx^+_j - \vy^\star_j\|^2 \;\le\; \frac{2\,\epsilon_{\gB}}{1/h - \kappa}.
\end{equation}
\end{lemma}

\begin{proof}
Each $f_j$ is $(1/h - \kappa)$-strongly convex with minimizer $\vy^\star_j$, so $f_j(\vy) - f_j(\vy^\star_j) \ge \frac{1}{2}(1/h - \kappa)\|\vy - \vy^\star_j\|^2$ for every $\vy$; take $\vy = \vx^+_j$, sum over $j$ and divide by $P$. The gap $\epsilon_{\gB}$ vanishes when the proximal points are themselves distinct candidates of $\gB$, and it is small when $\gB$ contains distinct candidates near them.
\end{proof}
\end{revisionarxivblock}

\subsection{Training}\label{app: training}

This section supports the training procedure of Section~\ref{sec: training}: Proposition~\ref{prop: what the field learns} characterizes what the velocity field learns, and Proposition~\ref{prop: straight paths} the pairing used there.

\begin{proposition}[What the velocity field learns]\label{prop: what the field learns}
Let the training tuples be $(\vx,\gC,\vx^0,\vx^+)$, where $\vx^0=\vx+\bm{\xi}$ with $\bm{\xi}\sim\mathcal{N}(0,\sigma_\xi^2 I)$ independent of $c:=(\vx,\gC)$ and the teacher's randomness, and $\vx^+$ denotes the targets obtained from (\ref{eq: assignment}). Let $\vx_\tau=(1-\tau)\vx^0+\tau\vx^+$, and let $v^\star$ minimize the flow-matching loss (\ref{eq: fm loss}) over measurable fields. Then:

\begin{enumerate}[label=(\roman*), leftmargin=2.2em, itemsep=0pt, topsep=2pt]
\item \textbf{Optimal velocity.} For each member $j$, the optimal field satisfies
\begin{equation}
v_j^\star(\vz,\tau;c)
=\E\big[\vx_j^+-\vx_j^0\mid\vx_\tau=\vz,\tau,c\big].
\end{equation}
If, as in the architecture of Section~\ref{sec: architecture}, the field for member $j$ depends on its own interpolated position and the context, but not on the transported positions of the other members, its optimal value is instead
\begin{equation}
\bar v_j^\star(\vz,\tau;c)
=\E\big[\vx_j^+-\vx_j^0
\mid\vx_{\tau,j}=\vz,\tau,c\big],
\end{equation}
which is the field characterized in Theorem~\ref{thm: attention}(i).
\item \textbf{Distributional transport.} If $\sigma_\xi>0$, the interpolated population law satisfies the continuity equation associated with $v^\star$. Under standard well-posedness assumptions, the flow starting from a fresh perturbation reproduces this law for every $\tau<1$, which converges weakly to the target law as $\tau\to1$. The same holds for each member's marginal law under $\bar v_j^\star$.
\item \textbf{Vanishing perturbation.} If $\sigma_\xi=0$, the optimal field along the training paths satisfies
\begin{equation}
v_j^\star(\vz,\tau;c)
=\frac{\vz_j-\vx_j}{\tau},
\qquad \tau>0.
\end{equation}
At $\tau=0$, the conditional-mean velocity is $\E[\vx_j^+\mid c]-\vx_j$. Selecting the differentiable trajectory with this initial velocity gives
\begin{equation}
\vz(\tau)=\vx_j+\tau\big(\E[\vx_j^+\mid c]-\vx_j\big),
\end{equation}
which ends at the regression $\E[\vx_j^+\mid c]$, regardless of the conditional target law. More generally, a deterministic flow from a single point cannot reproduce a nondegenerate target distribution.
\end{enumerate}
\end{proposition}

\begin{proof}
\emph{(i) Optimal velocity.}
For fixed $(\tau,c)$, the loss (\ref{eq: fm loss}) is the mean squared error of predicting the displacement $\vx^+-\vx^0$ from the interpolated population $\vx_\tau$. Its minimizer over measurable fields is therefore the conditional expectation
\begin{equation}
v^\star(\vz,\tau;c)
=\E[\vx^+-\vx^0\mid\vx_\tau=\vz,\tau,c],
\end{equation}
by the $L^2$ projection property. Taking the $j$-th component gives the first claim. If the field observes only $\vx_{\tau,j}$ and the context, the same argument gives the memberwise conditional expectation.

\emph{(ii) Distributional transport.}
Condition on $c$ and let $p_\tau$ denote the law of $\vx_\tau$. Differentiating the expectation of any smooth test function $\varphi$ along the straight interpolation gives
\begin{equation}
\begin{aligned}
\frac{\rd}{\rd\tau}\E[\varphi(\vx_\tau)\mid c]
&=\E\big[
\nabla\varphi(\vx_\tau)\cdot(\vx^+-\vx^0)
\mid c\big]\\
&=\E\big[
\nabla\varphi(\vx_\tau)\cdot
v^\star(\vx_\tau,\tau;c)\mid c\big],
\end{aligned}
\end{equation}
where the second equality follows from (i). This is the weak form of the continuity equation
\begin{equation}
\partial_\tau p_\tau
+\nabla\cdot(p_\tau v^\star)=0,
\end{equation}
as in standard flow matching \citep{lipman2023flow,tong2024improving_flow_with_ot}. Under the usual regularity conditions ensuring a well-defined flow, integrating $v^\star$ from the source law therefore reproduces $p_\tau$.

When $\sigma_\xi>0$, a fresh perturbation $\vx+\bm{\xi}$ has the same initial law as the training source. Moreover, $\vx_\tau\to\vx^+$ almost surely as $\tau\to1$, so $p_\tau$ converges weakly to the target law. Applying the same argument to a test function of $\vx_{\tau,j}$ alone establishes the corresponding result for the memberwise field $\bar v_j^\star$.

\emph{(iii) Vanishing perturbation.}
When $\sigma_\xi=0$, the source is fixed at $\vx^0=\vx$ given $c$. For any $\tau>0$, the interpolation
\begin{equation}
\vx_\tau=\vx+\tau(\vx^+-\vx)
\end{equation}
uniquely determines the target through
\begin{equation}
\vx^+=\vx+\frac{\vx_\tau-\vx}{\tau}.
\end{equation}
Consequently, the displacement is determined by the interpolated position, giving
\begin{equation}
v_j^\star(\vz,\tau;c)
=\frac{\vz_j-\vx_j}{\tau}
\end{equation}
along the training paths.

At $\tau=0$, all paths begin at $\vx$, so the conditional-mean velocity is
\begin{equation}
\E[\vx_j^+-\vx_j^0\mid c]
=\E[\vx_j^+\mid c]-\vx_j.
\end{equation}
The corresponding ODE $\dot{\vz}=(\vz-\vx_j)/\tau$ is singular at $\tau=0$: every trajectory $\vz(\tau)=\vx_j+\tau\vw$ solves it for $\tau>0$ and starts at $\vx_j$. If we prescribe the initial derivative $\dot{\vz}(0)=\E[\vx_j^+\mid c]-\vx_j$ and require a differentiable solution at $\tau=0$, this selects
\begin{equation}
\vz(\tau)=\vx_j+\tau\big(\E[\vx_j^+\mid c]-\vx_j\big),
\end{equation}
which ends at the regression $\E[\vx_j^+\mid c]$.

Thus, without perturbations, the natural conditional-mean initialization yields a regression trajectory rather than a distribution of targets. More fundamentally, a deterministic flow from the fixed source $\vx_j$ cannot reproduce a nondegenerate target law, motivating the Gaussian perturbation.
\end{proof}

\begin{proposition}[Straight paths under the assignment]\label{prop: straight paths}
Let $\pi$ solve (\ref{eq: assignment}) and $\vx_j^+=\vb_{\pi(j)}$. Then:

\begin{enumerate}[label=(\roman*), leftmargin=2.2em, itemsep=0pt, topsep=2pt]
\item \textbf{Optimal pairing.} The pairing $j\mapsto\vx_j^+$ is $W_2$-optimal between $\mu_{\vx^0}$ and $\mu_{\vx^+}$ for every $h$.
\item \textbf{Geodesic interpolation.} The straight interpolation
\begin{equation}
\mu_\tau:=\frac{1}{P}\sum_{j=1}^P
\delta_{(1-\tau)\vx_j^0+\tau\vx_j^+}
\end{equation}
is a constant-speed $W_2$ geodesic from $\mu_{\vx^0}$ to $\mu_{\vx^+}$.
\item \textbf{Non-crossing paths.} The paths $\tau\mapsto(1-\tau)\vx_j^0+\tau\vx_j^+$ of distinct members do not meet at any $\tau\in(0,1)$.
\end{enumerate}
\end{proposition}

\begin{proof}
\emph{(i) Optimal pairing.}
Let $S=\pi(\{1,\ldots,P\})$ be the set of selected candidates. For this fixed set, the assignment objective (\ref{eq: assignment}) is
\begin{equation}
\sum_{i\in S}J_i
+\frac{1}{2h}\sum_{j=1}^P
\|\vx_j^0-\vb_{\pi(j)}\|^2.
\end{equation}
The first term depends only on $S$, so the optimal assignment must minimize the total squared displacement among all bijections onto $S$. This is precisely the $W_2$-optimal matching between the two uniform empirical measures.

\emph{(ii) Geodesic interpolation.}
The linear interpolation of a $W_2$-optimal coupling is a constant-speed Wasserstein geodesic, known as displacement interpolation \citep[Thm.~7.2.2]{ambrosio2005gradient}. Applying this result to the optimal pairing from (i) gives the claim.

\emph{(iii) Non-crossing paths.}
Consider two distinct members $j\ne k$. Since swapping their assigned targets preserves feasibility, optimality of $\pi$ implies
\begin{equation}
\begin{aligned}
&\|\vx_j^0-\vx_j^+\|^2
+\|\vx_k^0-\vx_k^+\|^2\\
&\qquad\leq
\|\vx_j^0-\vx_k^+\|^2
+\|\vx_k^0-\vx_j^+\|^2.
\end{aligned}
\end{equation}
Expanding and simplifying gives
\begin{equation}
\langle\vx_j^0-\vx_k^0,\,
\vx_j^+-\vx_k^+\rangle\geq0.
\end{equation}
Now suppose their straight paths meet at some $\tau\in(0,1)$. Then
\begin{equation}
(1-\tau)(\vx_j^0-\vx_k^0)
=-\tau(\vx_j^+-\vx_k^+),
\end{equation}
which implies
\begin{equation}
\langle\vx_j^0-\vx_k^0,\,
\vx_j^+-\vx_k^+\rangle
=-\frac{\tau}{1-\tau}
\|\vx_j^+-\vx_k^+\|^2<0.
\end{equation}
The inequality is strict because the assignment selects distinct candidates, so $\vx_j^+\ne\vx_k^+$. This contradicts optimality of the pairing and proves that the paths cannot meet at any intermediate time.
\end{proof}

\subsection{Architecture}\label{app: arch}

\begin{proposition}[Covariance of the step]\label{prop: cov}
Let $\mathsf{T}(\vx^0,\gB,J;h)$ denote the assignment (\ref{eq: assignment}), and $\mathsf{T}_\lambda(\vx,\gB,J;\lambda,h)$ the soft step of Theorem~\ref{thm: attention}(ii). The following properties hold, with equalities for $\mathsf{T}$ understood as equalities of solution sets when the assignment is not unique.

\begin{enumerate}[label=(\roman*), leftmargin=2.2em, itemsep=0pt, topsep=2pt]
\item \textbf{Cost rescaling.} For $a>0$ and $b\in\R$,
\begin{equation}
\begin{aligned}
\mathsf{T}(\vx^0,\gB,aJ+b;h/a)
&=\mathsf{T}(\vx^0,\gB,J;h),\\
\mathsf{T}_\lambda(\vx,\gB,aJ+b;a\lambda,h/a)
&=\mathsf{T}_\lambda(\vx,\gB,J;\lambda,h).
\end{aligned}
\end{equation}
\item \textbf{Action-space transformations.} For $c>0$, orthogonal $Q$, and $\vm\in\R^d$, let $\Psi(\va):=cQ\va+\vm$. Then,
\begin{equation}
\begin{aligned}
&\mathsf{T}(\Psi(\vx^0),\Psi(\gB),J\circ\Psi^{-1};c^2h)\\
&\qquad=\Psi\big(\mathsf{T}(\vx^0,\gB,J;h)\big),
\end{aligned}
\end{equation}
and likewise for $\mathsf{T}_\lambda$, with parameters $(\lambda,c^2h)$.
\item \textbf{Permutation equivariance.} Both steps commute with relabelings of the population members and evaluated candidates.
\end{enumerate}
Consequently, with $\smash{\lambda=\tilde{\lambda}\sigma^k}$ and $\smash{h=\tilde{h}(s^k)^2/\sigma^k}$ as in Section~\ref{sec: architecture}, both steps and the heat substep are invariant to positive affine rescalings of the cost and equivariant to similarities of the action space, for fixed $(\tilde{\lambda},\tilde{h})$.
\end{proposition}

\begin{proof}
\emph{(i) Cost rescaling.}
Replacing $J$ by $aJ+b$ and $h$ by $h/a$ multiplies the assignment objective (\ref{eq: assignment}) by $a$ and adds the constant $Pb$. Since $a>0$, the minimizing assignments are unchanged.

For the soft step, additionally replacing $\lambda$ by $a\lambda$ changes the exponent of each weight to
\begin{equation}
-\frac{J_i+\|\vx-\vb_i\|^2/(2h)}{\lambda}
-\frac{b}{a\lambda}.
\end{equation}
The last term is independent of $i$ and cancels in the softmax, leaving the weights and their barycentric output unchanged.

\emph{(ii) Action-space transformations.}
Since $Q$ is orthogonal,
\begin{equation}
\|\Psi(\vx)-\Psi(\vb_i)\|^2
=c^2\|\vx-\vb_i\|^2.
\end{equation}
Thus, replacing $h$ by $c^2h$ preserves the displacement penalty. Moreover,
\begin{equation}
(J\circ\Psi^{-1})(\Psi(\vb_i))=J_i,
\end{equation}
so the candidate costs are unchanged. Hence, the assignment selects the same candidate indices, now at their transformed positions, proving the claim for $\mathsf{T}$.

The same argument preserves the soft weights $\alpha_i$. Since $\Psi$ is affine and $\sum_i\alpha_i=1$, its barycentric output satisfies
\begin{equation}
\sum_i\alpha_i\Psi(\vb_i)
=\Psi\left(\sum_i\alpha_i\vb_i\right),
\end{equation}
establishing the claim for $\mathsf{T}_\lambda$.

\emph{(iii) Permutation equivariance.}
Relabeling the members or candidates only reindexes the terms in the assignment objective, without changing the feasible matchings or their costs. Thus, the optimal assignments are relabeled accordingly. Similarly, permuting the candidates and their associated costs and reference weights permutes the softmax weights without changing the barycentric output. Both steps therefore commute with these relabelings.

Finally, the normalization in Section~\ref{sec: architecture} preserves these properties. Under $J\mapsto aJ+b$, the cost standard deviation scales as $\sigma^k\mapsto a\sigma^k$, so $(\lambda,h)\mapsto(a\lambda,h/a)$, as required by (i). Under $\Psi$, the population spread scales as $s^k\mapsto cs^k$, giving $h\mapsto c^2h$ while leaving $\lambda$ unchanged, as required by (ii). The Gaussian perturbation variance $\sigma_\xi^2=2\lambda h$ is therefore unchanged under cost rescaling and scales by $c^2$ under $\Psi$, establishing the corresponding properties of the heat substep.
\end{proof}

\begin{lemma}[A posterior mean over a finite set is a softmax attention]\label{lem: post}
Let $I \in \{1,\dots,M\}$ have prior $\pi \in \Delta^M$, let $\vm_i,\vb_i\in\R^d$, and let $\vu=\vm_I+s\,\bm{\varepsilon}$, where $s>0$ and $\bm{\varepsilon}\sim\mathcal{N}(0,I_d)$ is independent of $I$. Then,
\begin{equation}
\E[\vb_I\mid\vu]=\sum_i\alpha_i(\vu)\,\vb_i,\qquad
\alpha(\vu)=\operatorname{softmax}_i\Big(
\frac{\langle\vu,\vm_i\rangle}{s^2}+\beta_i
\Big),\qquad
\beta_i=\log\pi_i-\frac{\|\vm_i\|^2}{2s^2}.
\label{eq: post}
\end{equation}
Thus, the posterior mean takes the form of scaled dot-product attention with query $\vu$, keys $\vm_i$, values $\vb_i$, key biases $\beta_i$, and temperature $s^2$. Moreover, $\alpha(\vu)$ is the unique minimizer of
\begin{equation}
\min_{\alpha\in\Delta^M}
\sum_i\alpha_i\frac{\|\vu-\vm_i\|^2}{2s^2}
+\KL(\alpha\,\|\,\pi).
\end{equation}
\end{lemma}

\begin{proof}
By Bayes' rule, the posterior probability of candidate $i$ is
\begin{equation}
\alpha_i(\vu):=\Pr(I=i\mid\vu)
=\frac{\pi_i\exp\big(-\|\vu-\vm_i\|^2/(2s^2)\big)}
{\sum_l\pi_l\exp\big(-\|\vu-\vm_l\|^2/(2s^2)\big)}.
\end{equation}
Expanding the squared distance gives
\begin{equation}
-\frac{\|\vu-\vm_i\|^2}{2s^2}
=\frac{\langle\vu,\vm_i\rangle}{s^2}
-\frac{\|\vm_i\|^2}{2s^2}
-\frac{\|\vu\|^2}{2s^2}.
\end{equation}
The last term is independent of $i$ and cancels in the normalization, yielding the softmax weights in (\ref{eq: post}). The posterior mean follows by averaging the values $\vb_i$ under these probabilities.

For the variational characterization, let $c_i=\|\vu-\vm_i\|^2/(2s^2)$ and $Z=\sum_i\pi_i e^{-c_i}$. Since $\alpha_i(\vu)=\pi_i e^{-c_i}/Z$, we have
\begin{equation}
\sum_i\alpha_i c_i+\KL(\alpha\,\|\,\pi)
=\KL(\alpha\,\|\,\alpha(\vu))-\log Z.
\end{equation}
The KL divergence is nonnegative and vanishes only when $\alpha=\alpha(\vu)$, proving that the posterior weights are the unique minimizer.
\end{proof}

\begin{proof}[Proof of Theorem~\ref{thm: attention}(i)]
Fix a population member $j$ and condition on the context $c$ and the teacher's evaluated set $\gB$. Since $\gB$ is generated independently of the perturbations $\bm{\xi}$, the source $\vx_j^0$ has density $p_0=\mathcal{N}(\vx_j,\sigma_\xi^2 I)$. Let
\begin{equation}
g_i(\vu):=\Pr(\vx_j^+=\vb_i\mid\vx_j^0=\vu,c,\gB)
\end{equation}
denote the probability of assigning member $j$ to candidate $\vb_i$, given its source position. We suppress the conditioning on $(c,\gB)$ below.

For $\tau<1$, the straight interpolation $\vx_{\tau,j}=(1-\tau)\vx_j^0+\tau\vx_j^+$ implies that, if the target is $\vb_i$ and the interpolated position is $\vz$, the source must be
\begin{equation}
\vu_i(\vz,\tau):=\frac{\vz-\tau\vb_i}{1-\tau}.
\end{equation}
Since this transformation has Jacobian determinant $(1-\tau)^d$, the joint density of $\vx_{\tau,j}=\vz$ and $\vx_j^+=\vb_i$ is $(1-\tau)^{-d}p_0(\vu_i)g_i(\vu_i)$. Bayes' rule therefore gives
\begin{equation}
\alpha_i(\vz,\tau)
:=\Pr(\vx_j^+=\vb_i\mid\vx_{\tau,j}=\vz)
=\frac{p_0(\vu_i)g_i(\vu_i)}
{\sum_l p_0(\vu_l)g_l(\vu_l)}.
\end{equation}
Moreover, the straight path satisfies $\vx_j^+-\vx_j^0=(\vx_j^+-\vx_{\tau,j})/(1-\tau)$. Thus, by Proposition~\ref{prop: what the field learns}(i), the optimal velocity field is
\begin{equation}
\begin{aligned}
v^\star(\vz,\tau)
&=\E[\vx_j^+-\vx_j^0\mid\vx_{\tau,j}=\vz]\\
&=\frac{1}{1-\tau}
\left(\sum_i\alpha_i(\vz,\tau)\vb_i-\vz\right).
\end{aligned}
\end{equation}

We next identify the attention weights. Since $p_0$ is Gaussian, expanding the squared distance gives
\begin{equation}
\begin{aligned}
\log p_0(\vu_i)
&=-\frac{\|\vz-\tau\vb_i-(1-\tau)\vx_j\|^2}
{2(1-\tau)^2\sigma_\xi^2}+\mathrm{const}\\
&=\frac{\tau\langle\vz-(1-\tau)\vx_j,\vb_i\rangle}
{(1-\tau)^2\sigma_\xi^2}
-\frac{\tau^2\|\vb_i\|^2}
{2(1-\tau)^2\sigma_\xi^2}+\mathrm{const},
\end{aligned}
\end{equation}
where the constant is independent of $i$. Substituting into the posterior weights yields
\begin{equation}
\alpha_i(\vz,\tau)
=\operatorname{softmax}_i\left(
\langle\vq,\vb_i\rangle+\beta_i+\log g_i(\vu_i)
\right),
\end{equation}
with
\begin{equation}
\vq=\frac{\tau[\vz-(1-\tau)\vx_j]}
{(1-\tau)^2\sigma_\xi^2},
\qquad
\beta_i=-\frac{\tau^2\|\vb_i\|^2}
{2(1-\tau)^2\sigma_\xi^2},
\end{equation}
which establishes (\ref{eq: posterior_field}). When $g_i(\vu)=w_i$ is independent of the source, $\log w_i$ becomes a fixed key bias, and the weights reduce to the dot-product attention of Lemma~\ref{lem: post} with prior $w$, keys $\vm_i=(1-\tau)\vx_j+\tau\vb_i$, and $s=(1-\tau)\sigma_\xi$. The resulting logits coincide with those above up to a term independent of $i$, recovering the independent-coupling case of \citet{bertrand2026closed}.

Finally, marginalizing over the teacher's randomness gives the optimal field conditioned only on the deployment context:
\begin{equation}
v^\star(\vz,\tau;c)
=\E\!\left[
v^\star(\vz,\tau;c,\gB)
\mid \vx_{\tau,j}=\vz,c
\right],
\end{equation}
by the tower property of conditional expectation.
\end{proof}

\paragraph{The relaxed step.}
To obtain a soft version of the proximal assignment, consider couplings $\gamma\in\R_+^{P\times M}$ with row sums $1/P$, where $\gamma_{ji}$ represents the mass transported from member $j$ to candidate $\vb_i$. Let $\nu_0$ be a strictly positive reference distribution on $\gB$. Rather than regularizing the target marginal $\gamma^\top\bm{1}$, we regularize the coupling itself through $\lambda\KL(\gamma\,\|\,\frac{1}{P}\bm{1}\otimes\nu_0)$. By the KL chain rule,
\begin{equation}
\KL\Big(\gamma\,\Big\|\,\frac{1}{P}\bm{1}\otimes\nu_0\Big)
=
\KL(\gamma^\top\bm{1}\,\|\,\nu_0)+I(\gamma),
\end{equation}
where $I(\gamma)\geq0$ is the mutual information between source members and destinations. Thus, regularizing the coupling gives an upper bound on the corresponding marginal-regularized objective, with an additional penalty $\lambda I(\gamma)$. The bound is tight for product couplings, as occurs when $h\to\infty$.

The advantage is that the resulting problem separates over population members. Writing $\gamma_{ji}=\alpha_{ji}/P$, each member $j$ independently solves
\begin{equation}
\min_{\alpha\in\Delta^M}\ F_{\vx_j}(\alpha)
:=
\sum_i\alpha_i\Big[
J_i+\frac{\|\vx_j-\vb_i\|^2}{2h}
\Big]
+\lambda\KL(\alpha\,\|\,\nu_0).
\label{eq: relaxed}
\end{equation}
Its barycentric output $\sum_i\alpha_i\vb_i$ is the soft proximal step of Theorem~\ref{thm: attention}(ii). Unlike the assignment (\ref{eq: assignment}), this relaxation allows different members to select the same candidate; the original assignment is recovered in the zero-temperature limit only when the distinctness constraints are retained. Finally, since $\nu_0$ is supported on the finite set $\gB$, this is a discrete surrogate, while the continuum entropy of (\ref{eq: free energy}) is handled separately by the heat substep.

\begin{proof}[Proof of Theorem~\ref{thm: attention}(ii)]
For a fixed source $\vx$, let
\begin{equation}
c_i:=J_i+\frac{\|\vx-\vb_i\|^2}{2h},
\qquad
Z:=\sum_i\nu_{0,i}e^{-c_i/\lambda}.
\end{equation}
The objective in (\ref{eq: relaxed}) can be written as
\begin{equation}
\begin{aligned}
F_{\vx}(\alpha)
&=\sum_i\alpha_i c_i+\lambda\KL(\alpha\,\|\,\nu_0)\\
&=\lambda\KL(\alpha\,\|\,\alpha^\star)-\lambda\log Z,
\end{aligned}
\end{equation}
where $\alpha_i^\star=\nu_{0,i}e^{-c_i/\lambda}/Z$. Since the KL divergence is nonnegative and vanishes only when $\alpha=\alpha^\star$, this is the unique minimizer.

Expanding the squared distance gives
\begin{equation}
-\frac{\|\vx-\vb_i\|^2}{2h\lambda}
=\frac{\langle\vx,\vb_i\rangle}{h\lambda}
-\frac{\|\vb_i\|^2}{2h\lambda}
+\mathrm{const},
\end{equation}
where the constant is independent of $i$. Thus,
\begin{equation}
\alpha_i^\star(\vx)
=\operatorname{softmax}_i\left(
\frac{\langle\vx,\vb_i\rangle}{h\lambda}+\beta_i
\right),
\qquad
\beta_i=\log\nu_{0,i}-\frac{J_i}{\lambda}
-\frac{\|\vb_i\|^2}{2h\lambda}.
\end{equation}
The barycentric output $\sum_i\alpha_i^\star(\vx)\vb_i$ is therefore a dot-product attention operation with query $\vx$, keys and values $\vb_i$, and additive key biases $\beta_i$.

This proves the claim. Two further observations are used in Remark~\ref{rem: converse}. First, to realize the Euclidean transport term using projected attention with logits $\langle W_q\vx,W_k\vb_i\rangle/\sqrt{d_h}$, the projections must satisfy
\begin{equation}
W_q^\top W_k=\frac{\sqrt{d_h}}{h\lambda}I_d.
\end{equation}
Since the LHS has rank at most $d_h$, this is possible for arbitrary $\vx,\vb_i\in\R^d$ if and only if $d_h\geq d$.

Second, conversely, consider any projected attention layer with logits
\begin{equation}
r_i(\vx)=\frac{\langle W_q\vx,W_k\vb_i\rangle}{\sqrt{d_h}}+\beta_i
=\vx^\top A\vb_i+\beta_i,
\qquad
A:=\frac{W_q^\top W_k}{\sqrt{d_h}}.
\end{equation}
Setting $p_i\propto e^{\beta_i}$, its softmax weights are the unique minimizer of
\begin{equation}
\min_{\alpha\in\Delta^M}
\sum_i\alpha_i(-\vx^\top A\vb_i)
+\KL(\alpha\,\|\,p),
\end{equation}
by the same KL argument above. Thus, projected attention corresponds to an entropy-regularized step with bilinear ground cost $-\vx^\top A\vb_i$, where $\operatorname{rank}(A)\leq d_h$. When $A$ is symmetric positive semidefinite, this becomes the squared Mahalanobis distance $\|\vx-\vb_i\|_A^2/2$, with key biases $\beta_i=\log p_i-\vb_i^\top A\vb_i/2$ for a prescribed reference distribution $p$, up to a common additive constant.
\end{proof}

\begin{remark}[Converse]\label{rem: converse}
Theorem~\ref{thm: attention}(ii) also admits a converse interpretation. For any projections $W_q,W_k$ and key biases $\beta_i$, the cross-attention weights
\begin{equation}
\alpha_i(\vx)=\operatorname{softmax}_i\left(
\frac{\langle W_q\vx,W_k\vb_i\rangle}{\sqrt{d_h}}+\beta_i
\right)
\end{equation}
uniquely minimize
\begin{equation}
\sum_i\alpha_i(-\vx^\top A\vb_i)
+\KL(\alpha\,\|\,p),
\end{equation}
where $A:=W_q^\top W_k/\sqrt{d_h}$ and $p_i\propto e^{\beta_i}$. Thus, dot-product attention corresponds to a KL-regularized step with bilinear ground cost $-\vx^\top A\vb_i$, where $\operatorname{rank}(A)\leq d_h$. For a fixed reference distribution $p$, this becomes a squared Mahalanobis step when $A$ is symmetric positive semidefinite and $\beta_i=\log p_i-\vb_i^\top A\vb_i/2$, up to an additive constant. In particular, representing the full-dimensional Euclidean step of (\ref{eq: soft step}) requires $d_h\geq d$.
\end{remark}

\begin{proof}[Proof of Corollary~\ref{cor: corners}]
We first connect the attention weights to Theorem~\ref{thm: attention}(ii), then construct the MPPI and CEM updates.

From Theorem~\ref{thm: attention}(ii), the attention logits are
\begin{equation}
r_i(\vx)=\frac{\langle\vx,\vb_i\rangle}{h\lambda}
+\log\nu_{0,i}-\frac{J_i}{\lambda}
-\frac{\|\vb_i\|^2}{2h\lambda}.
\end{equation}
As $h\to\infty$, the transport-dependent terms vanish, leaving weights
\begin{equation}
w_i=\frac{\nu_{0,i}e^{-J_i/\lambda}}
{\sum_l\nu_{0,l}e^{-J_l/\lambda}},
\end{equation}
which are independent of the querying member. For uniform $\nu_0$, these are precisely the MPPI weights, obtained by setting the query-independent attention logits to $\beta_i=-J_i/\lambda$. For CEM, instead choose $\beta_i=0$ for $i\in\gE$ and $\beta_i=-\kappa$ otherwise, where $\gE$ is the elite set. As $\kappa\to\infty$, the resulting weights converge to $w_i=\mathbb{1}\{i\in\gE\}/|\gE|$. Both patterns can be represented using the cost and rank features of the evaluated population, with queries that are constant across members.

It remains to show how these weights recover the population updates. Let
\begin{equation}
\bm{\mu}=\frac1P\sum_{i=1}^P\vx_i,
\qquad
\bm{\mu}_w=\sum_{i=1}^P w_i\vx_i
\end{equation}
be the ordinary and weighted population means. Choose two attention heads with values $\vx_i$: the first uses the weights $w_i$ and returns $\bm{\mu}_w$, while the second uses uniform weights $1/P$ and returns $\bm{\mu}$. The output projection takes their difference, giving the same velocity to every member,
\begin{equation}
v_j=\bm{\mu}_w-\bm{\mu}.
\end{equation}
Since this velocity is independent of the member's position and flow time, one Euler step of unit length gives
\begin{equation}
\vx_j^+=\vx_j+\bm{\mu}_w-\bm{\mu},
\qquad
\frac1P\sum_{j=1}^P\vx_j^+
=\bm{\mu}_w=\sum_{i=1}^P w_i\vx_i.
\end{equation}
Thus, the updated population mean equals the MPPI weighted mean, or converges to the CEM elite mean as $\kappa\to\infty$.

Finally, the same construction applies to the normalized coordinates used by our architecture. With $\hat{\vx}_i=(\vx_i-\bar{\vx})/s$, the attention output in normalized coordinates satisfies
\begin{equation}
s\left(\sum_i w_i\hat{\vx}_i
-\frac1P\sum_i\hat{\vx}_i\right)
=\sum_i w_i\vx_i-\frac1P\sum_i\vx_i.
\end{equation}
Hence, rescaling the output recovers exactly the same displacement in the original action space.
\end{proof}

\section{Background on Flow Matching}\label{app: flow matching}

Flow-based generative models \citep{chen2018neural, lipman2023flow} transport a base distribution $p_0$ to a target $p_1$ on $\R^D$. The transport is defined by a learnable velocity field $v_\theta : \R^D \times [0,1] \to \R^D$ which induces the ordinary differential equation (ODE):
\begin{equation}
\dot{\vx}(\tau) = v_\theta\big(\vx(\tau), \tau \big), \qquad \vx(0) \sim p_0,
\label{eq:fm_ode}
\end{equation}
whose time-one map pushes $p_0$ forward to $p_1$. 
Flow matching \citep{lipman2023flow, liu2023flow, albergo2023building} trains $v_\theta$ through direct regression, without integrating (\ref{eq:fm_ode}) during training. Sampling a coupling $(\vx_0, \vx_1) \sim \pi$ with marginals $p_0$ and $p_1$ (e.g., the independent one $\pi = p_0 \otimes p_1$) and interpolating linearly, $\vx_\tau = (1-\tau)\, \vx_0 + \tau\, \vx_1$, the conditional flow matching loss
\begin{equation}
\mathcal{L}_{\mathrm{CFM}}(\theta) = \E_{\tau \sim \mathrm{Unif}[0,1],\, (\vx_0, \vx_1) \sim \pi} \Big[ \big\| v_\theta(\vx_\tau, \tau) - (\vx_1 - \vx_0) \big\|^2 \Big]
\label{eq:cfm_loss}
\end{equation}
is minimized by $v^\star(\vx,\tau) = \E\big[\vx_1 - \vx_0 \mid \vx_\tau = \vx\big]$, the velocity field that generates the marginals $p_\tau$ of $\vx_\tau$ and hence transports $p_0$ to $p_1$ for any such $\pi$ \citep{tong2024improving_flow_with_ot}.
Inference then amounts to integrating (\ref{eq:fm_ode}), typically with a few Euler steps. Conditional generation is obtained by providing context $c$ to the velocity field, $v_\theta(\vx, \tau, c)$, trained on triplets $(\vx_0, \vx_1, c)$ drawn jointly.

\section{FlowOpt Policy Architecture Details}
\label{app:architecture}

\subsection{Rollout and context construction}
\label{app:context}

Given the current state $\hat{\vs}_0 \in \R^{n_s}$ and the incoming population $\gA^k = \{\bar{\va}_j^k\}_{j=1}^P$, all members are rolled out in parallel under the dynamics and scored under the cost,
\begin{equation}
\vs_{j,0} = \hat{\vs}_0, \qquad \vs_{j,t+1} = f(\vs_{j,t},\, \va^k_{j,t}), \qquad
J^k_j = \sum_{t=0}^{H-1} \ell(\vs_{j,t},\, \va^k_{j,t}) + \phi(\vs_{j,H}),
\end{equation}
for $t = 0, \dots, H-1$. The context of iteration $k$ is given by
\begin{equation}
\gC^k = \big\{ (\rho^k_j, \tilde{J}^k_j ) \big\}_{j=1}^P \in [0,1]^P \times \R^P,
\end{equation}
and consists of the normalized rank and standardized cost of every member as defined in Appendix~\ref{app:norm}. The context is computed once per outer iteration $k$, at the cost of $P$ parallel rollouts of $H$ steps each, and held fixed during the flow integration of that iteration.

\subsection{Normalization}
\label{app:norm}
All cost quantities enter the network only through statistics that are invariant to positive affine rescaling of the cost. The member-level total-cost channel and rank channel are
\begin{equation}
\tilde{J}_j = \frac{J_j - \mu^J}{\sigma^J + \varepsilon},
\qquad
\rho_j = \frac{\mathrm{rank}(J_j) - 1}{P - 1} \in [0, 1],
\label{eq:app-zJ}
\end{equation}
with $\mu^J, \sigma^J$ computed across members and $\mathrm{rank}(J_j) \in \{1, \dots, P\}$ the rank of $J_j$ within the population (ascending).

The actions are normalized in the same spirit, relative to the evaluated population $\vx^k$ at $\tau=0$. Let
\begin{equation}
\bar{\vx}^k = \frac{1}{P}\sum_{j=1}^P \vx^k_j,
\qquad
s^k = \sqrt{\frac{1}{P H n_a}\sum_{j=1}^P \big\|\vx^k_j - \bar{\vx}^k\big\|^2} + \varepsilon,
\label{eq:app-action-stats}
\end{equation}
denote the population mean and the root-mean-square deviation of its entries, both computed once per outer iteration and held fixed during the flow. Every member, evaluated or transported, is presented to the network in standardized coordinates $\smash{\hat{\vx}_j = (\vx_j - \bar{\vx}^k)/s^k}$, and the network output $g_\theta$ is rescaled to action units, $\smash{v_\theta = s^k g_\theta}$. 
As a result, $v_\theta(\alpha\vx, \tau; \alpha\vx^k, \gC^k) = \alpha\, v_\theta(\vx, \tau; \vx^k, \gC^k)$ for $\alpha > 0$, so displacements are measured in units of the population's own spread.

\subsection{Attention Mechanism}
\label{app:blocks}
The network has a single cross-attention layer, which is the only place members communicate. Let $\smash{c_j = (\rho_j, \tilde{J}_j) \in \R^2}$ denote the context of member $j$ and let $\smash{W_x: \R^{H n_a} \to \R^d}$ and $\smash{W_c: \R^2 \to \R^d}$ be linear embeddings, with $d = 192$. The transported member $i$ at flow position $\hat{\vx}_i(\tau)$ forms a query token, and every evaluated member $j$ of the population forms a key/value token,
\begin{equation}
q_i = \mathrm{LN}_1\big(W_q\,(W_x \hat{\vx}_i(\tau) + W_c c_i)\big), \qquad
k_j = \mathrm{LN}_1\big([\,W_x \hat{\vx}^k_j\, ;\; W_c c_j\,]\big), \qquad q_i, k_j \in \R^{2d},
\end{equation}
where $[\,\cdot\,;\,\cdot\,]$ is concatenation and $W_q: \R^{d} \to \R^{2d}$, so that all tokens have width $2d = 384$. The key/value tokens are computed once per iteration from the evaluated population and held fixed during the flow integration. Each transported member $i$ carries the context $c_i$ of its own evaluated copy, which serves as a label of its origin rather than as an estimate of its cost at the current position. Multi-head attention with $n_h = 4$ heads of width $d_h = 2d/n_h = 96$ then computes
\begin{equation}
\alpha^h_{ij} = \mathrm{softmax}_j\!\left( \frac{\langle W^h_Q q_i,\, W^h_K k_j \rangle}{\sqrt{d_h}} \right), \quad
o_i = W_O \Big[ \textstyle\sum_{j=1}^P \alpha^h_{ij}\, W^h_V k_j \Big]_{h=1}^{n_h}, \quad
a_i = \mathrm{LN}_2(q_i + o_i).
\end{equation}
Each head forms a convex combination of the population tokens, with weights depending on the positions and costs of both $i$ and $j$; the residual retains member $i$'s own token, so the head below can express a displacement \emph{from} $\hat{\vx}_i$. Since no positional encoding is used and the statistics of Appendix~\ref{app:norm} are symmetric in $j$, the layer is permutation-equivariant over members.

\subsection{Velocity head}
\label{app:head}
The attended token is concatenated with the flow-time embedding of width $d$,
\begin{equation}
h_i = \big[\, a_i \,;\; W_\tau \tau + b_\tau \,\big] \in \R^{3d},
\end{equation}
and passed through a member-wise MLP with three hidden layers of width $m = 512$,
\begin{subequations}
\begin{align}
g_\theta(\hat{\vx}_i, \tau) &= \mathrm{MLP}(h_i) \in \R^{H n_a}, \\
\mathrm{MLP} &= \mathrm{Lin}(3d, m) \to \mathrm{LN} \to \mathrm{ELU} \to [\mathrm{Lin}(m,m) \to \mathrm{ELU}] \times 2 \to \mathrm{Lin}(m, H n_a).
\end{align}
\end{subequations}
The output layer is initialized to zero, so that the untrained field leaves the population unchanged. The velocity in action units is $v_\theta = s^k g_\theta$, as in Appendix~\ref{app:norm}. The head acts on each member independently; all cross-member information enters through $a_i$. In summary, the field computes
\begin{equation}
v_\theta(\vx_i, \tau) = s^k\, g_\theta\Big( \hat{\vx}_i,\ c_i,\ \textstyle\sum_j \alpha_{ij}\, V(\hat{\vx}^k_j, c_j),\ \tau \Big),
\end{equation}
i.e.\ a member's own position together with a cost-weighted summary of the population, the same information a CEM or MPPI update consumes, with the weighting and the step size learned.

\section{Experiment Details}
\label{sec: appendix experiment details}

\subsection{Environments and Tasks}

\subsubsection{CarNav environment}
\label{app:carnav}

CarNav is a custom planar navigation task with a bicycle vehicle, obstacles and a goal that moves during the episode. A planner must handle a non-holonomic constraint, obstacle avoidance with a finite body, repeated re-targeting, and a multimodal landscape with several local minima.

\paragraph{Dynamics.}
The vehicle is a kinematic bicycle with the rear axle as reference point. The state $s = (p_x, p_y, \theta, v)$ includes the position, heading and forward speed, and the action $a = (\alpha, \delta)$ the longitudinal acceleration and steering angle. The dynamics are given by
\begin{equation}
\dot p_x = v\cos\theta,\qquad
\dot p_y = v\sin\theta,\qquad
\dot\theta = \tfrac{v}{L}\tan\delta,\qquad
\dot v = \alpha,
\end{equation}
with wheelbase $L = 1\,\mathrm{m}$, integrated by forward Euler at $\Delta t = 0.05\,\mathrm{s}$. The steering angle is clamped to $|\delta| \le \delta_{\max} = 0.6\,\mathrm{rad}$ inside the dynamics, so the turning radius is at least $L/\tan\delta_{\max} \approx 1.5\,\mathrm{m}$ and the heading cannot change at zero speed. The acceleration is unbounded but regularized by the cost. The vehicle occupies a disk of radius $r_{\mathrm{car}} = 0.5\,\mathrm{m}$ around its reference point.

\paragraph{Cost.}
The objective is given by
\begin{equation}
\begin{aligned}
J(s_{0:H}, a_{0:H-1}) ={}&
  w_{\mathrm{run}} \sum_{t=0}^{H-1} \|p_t - g_p\|^2
  + \sum_{t=0}^{H-1} \big( w_\alpha\, \alpha_t^2 + w_\delta\, \delta_t^2 \big) \\
&+ w_{\mathrm{term}} \Big( \|p_H - g_p\|^2
  + w_\theta\, \mathrm{wrap}(\theta_H - g_\theta)^2
  + w_v\, v_H^2 \Big)
  \\
&  + w_{\mathrm{obs}} \sum_{t=0}^{H} \sum_{i} \max\!\big(0,\; \rho_i - \|p_t - c_i\|\big)^2 ,
\end{aligned}
\end{equation}
where $g_p$ and $g_\theta$ are the goal position and heading, $\mathrm{wrap}(\cdot)$ maps an angle difference to $(-\pi, \pi]$, and $\rho_i = r_i + r_{\mathrm{car}} + m$ is the exclusion radius of obstacle $i$ with center $c_i$, physical radius $r_i$ and safety margin $m = 0.1\,\mathrm{m}$ (a Minkowski inflation of the obstacle by the vehicle body). The weights are $w_{\mathrm{run}} = 1$, $w_\alpha = 0.5$, $w_\delta = 2$, $w_{\mathrm{term}} = 5$, $w_\theta = 1$, $w_v = 1$, and $w_{\mathrm{obs}} = 5000$.

\paragraph{Problem generation.}
The start position, heading and speed are uniform in $[-4, 4]^2\,\mathrm{m}$, $[-\pi, \pi]$ and $[-0.5, 0.5]\,\mathrm{m/s}$, respectively. The first goal position is uniform in $[-6, 6]^2\,\mathrm{m}$, redrawn until it is at least $6.5\,\mathrm{m}$ from the start, and the goal heading is uniform in $[-\pi, \pi]$. Three circular obstacles with radii uniform in $[0.3, 0.8]\,\mathrm{m}$ are placed along the start-to-goal segment with a lateral offset uniform in $[-0.5, 0.5]\,\mathrm{m}$. Obstacle centers are kept at least $2.4\,\mathrm{m}$ from both the start and the goal, which leaves $1\,\mathrm{m}$ of turning room, and do not overlap. The two subsequent goals of the episode are uniform in $[-6, 6]^2\,\mathrm{m}$ with uniform headings, rejected if they lie within the exclusion radius plus $1\,\mathrm{m}$ of turning room of any obstacle. The obstacles stay fixed throughout.

\paragraph{Closed-loop evaluation.}
All planners run in receding horizon with $H = 60$ ($3\,\mathrm{s}$ of look-ahead), re-planning at every step and warm-starting from the previous plan shifted by one step. An episode lasts $300$ closed-loop steps ($15\,\mathrm{s}$) and the goal switches every $100$ steps ($5\,\mathrm{s}$), giving three goal segments. The executed action is the planner's nominal. The reported closed-loop cost sums the cost above over the three goal segments, each evaluated on its executed trajectory under its own goal, with the terminal terms applied at its last step. The $100$ held-out evaluation instances are drawn with a fixed seed and shared by all methods.

\subsubsection{HalfCheetah environment}
\label{app:cheetah}

HalfCheetah is the standard planar MuJoCo running task, simulated from the unmodified Gym model with MuJoCo Warp for GPU-batched rollouts. The objective is a running gait with no goal or obstacles, and the difficulty lies in the contact dynamics and the coordination of six actuators.

\paragraph{Dynamics.}
The state is $s = (q, \dot q) \in \mathbb{R}^{18}$, where $q = (p_x, p_z, \phi, q_{1:6})$ includes the horizontal and vertical position of the torso, its pitch angle, and the six leg joint angles (back thigh, shin, foot; front thigh, shin, foot). The action $a \in [-1, 1]^6$ is the normalized motor command of the six joints, clipped to $[-1, 1]$ and scaled by the model's gear ratios into torques. The planner acts at $\Delta t = 0.05\,\mathrm{s}$ with two semi-implicit Euler substeps of $0.025\,\mathrm{s}$ each, and the action is held constant within a step.

\paragraph{Cost.}
The objective is given by
\begin{equation}
\begin{aligned}
J(s_{1:H}, a_{0:H-1}) &=
  \frac{1}{H}\sum_{t=1}^{H}\Big(
  - w_{\mathrm{fwd}}\, \dot p_{x,t}
  + w_{\mathrm{alive}} \max(0,\; z_{\min} - p_{z,t})^2
  + w_{\mathrm{pitch}}\, \phi_t^2 \Big)
  \\
& \quad + \frac{w_a}{H}\sum_{t=0}^{H-1} \tfrac{1}{6}\|a_t\|^2 ,
\end{aligned}
\end{equation}
with $w_{\mathrm{fwd}} = 1$, $w_{\mathrm{alive}} = 5$, $z_{\min} = -0.3\,\mathrm{m}$ (relative to the torso's rest height), $w_{\mathrm{pitch}} = 3$ and $w_a = 0.1$. The first term rewards forward speed, the second penalizes the torso dropping below a height floor, the third penalizes flipping and the last penalizes effort, all averaged over the horizon. We report $R = -J$ as a return (higher is better).

\paragraph{Problem generation.}
A problem instance corresponds to an initial state, drawn by perturbing the default standing pose with Gaussian noise of standard deviation $0.05\,\mathrm{m}$ on the torso height, $0.2\,\mathrm{rad}$ on the pitch, $0.1\,\mathrm{rad}$ on each joint angle and $0.5$ on every generalized velocity, with zero horizontal position. The planner must thus recover from off-balance starts and accelerate into a gait. The $64$ held-out evaluation starts are drawn with a fixed seed and shared by all methods.

\paragraph{Closed-loop evaluation.}
All planners run in receding horizon with $H = 20$ ($1\,\mathrm{s}$ of look-ahead), re-planning at every step and warm-starting from the previous plan shifted by one step. The executed action is the planner's nominal. The reported closed-loop return is $R = -J$ of the executed trajectory over an episode of $100$ steps ($5\,\mathrm{s}$) under the cost above, i.e., the average forward speed over the episode minus the average height, pitch, and action penalties.

\subsubsection{Push-T environment}
\label{app:pusht}

Push-T is the planar non-prehensile manipulation task introduced by \citet{chi2025diffusionpolicy}, in which a point pusher must push a T-shaped block across a floor to a fixed goal pose. Since contact is intermittent and unilateral, the pusher must go around the block and push from different sides, which makes the cost landscape multimodal and poorly served by a single-mode Gaussian planner.

\paragraph{Dynamics.}
The scene is simulated in MuJoCo (via MuJoCo Warp for GPU-batched rollouts) with a $0.01\,\mathrm{s}$ timestep and a sliding friction coefficient of $1.0$ on every surface. The T-block is the union of a $2 \times 8\,\mathrm{cm}$ bar and a $6 \times 2\,\mathrm{cm}$ crossbar, with three planar degrees of freedom, and the pusher is a sphere of radius $1\,\mathrm{cm}$ on two planar slide joints driven by velocity servos. The state is $\smash{s = (b_x, b_y, b_\theta, q_x, q_y, \dot b_x, \dot b_y, \dot b_\theta,
\dot q_x, \dot q_y) \in \mathbb{R}^{10}}$, the block pose and pusher position with their rates. The action $a = (v_x, v_y)$ is the commanded pusher velocity, clipped to $\smash{[-v_{\max}, v_{\max}]^2}$ with $v_{\max} = 0.5\,\mathrm{m/s}$. The planner's action knots are spaced $\Delta t = 0.1\,\mathrm{s}$ apart ($10$ simulator substeps), with the command held constant within a knot, so a knot displaces the pusher by at most $5\,\mathrm{cm}$.

\paragraph{Cost.}
The goal block pose is $g = (g_p, g_\theta) = (0, 0, 0)$, and the per-state cost is
\begin{equation}
c(s) = w_{\mathrm{pos}} \|b - g_p\|^2
     + w_{\mathrm{ang}}\, \mathrm{wrap}(b_\theta - g_\theta)^2
     + w_{\mathrm{close}} \|b - q\|^2 ,
\end{equation}
where $b = (b_x, b_y)$ is the block position, $q = (q_x, q_y)$ the pusher position, and $\mathrm{wrap}(\cdot)$ maps an angle difference to $(-\pi, \pi]$. The planning objective over a horizon of $H$ knots is
\begin{equation}
J(s_{0:H}, a_{0:H-1}) = \Delta t \sum_{t=0}^{H-1} c(s_t) + c(s_H),
\end{equation}
with $w_{\mathrm{pos}} = 1$, $w_{\mathrm{ang}} = 1$, and $w_{\mathrm{close}} = 0.01$. The first two terms drive the block to the goal pose, and the small third term keeps the pusher close enough to the block to act on it.

\paragraph{Problem generation.}
A problem instance corresponds to an initial state with zero velocities. The block's offset from the goal is uniform in $[-0.2, 0.2]^2\,\mathrm{m}$ with yaw in $[-\pi, \pi]$, and the pusher's offset from its home position, $10\,\mathrm{cm}$ in front of the goal, is uniform in $[-0.2, 0.2]^2\,\mathrm{m}$, so it may start in contact with, or inside, the block. All methods share $192$ evaluation starts ($3$ seeds of $64$).

\paragraph{Closed-loop evaluation.}
All planners run in receding horizon with $H = 20$ knots ($2\,\mathrm{s}$ of look-ahead) over episodes of $100$ closed-loop steps ($10\,\mathrm{s}$), re-planning at every knot and warm-starting from the previous plan shifted by one knot. Since the population mean of a sampling-based planner averages over incompatible pushing strategies and fixes the block's angle without displacing it, every sampler executes the cheapest sample of its final population instead. An episode is a success if, at its final step, the block position is within $2\,\mathrm{cm}$ of the goal and its yaw within $0.1\,\mathrm{rad}$ ($5.7^\circ$).

\subsubsection{Walker environment}
\label{app:walker}

Walker is the standard MuJoCo planar biped, which has to get up from a random, often fallen, configuration and walk at a target speed while keeping its torso upright at a target height. Since the cost describes no gait, walking has to emerge from the planner as the cheapest way to hold both with bounded torques, and the short planning horizon makes this a hard receding-horizon problem.

\paragraph{Dynamics.}
The scene is simulated with MuJoCo Warp at a $0.01\,\mathrm{s}$ timestep. The robot is a torso on a planar root with two legs of hip, knee and ankle hinges and a floor friction coefficient of $0.5$. The hip, knee and ankle ranges are $[-20^\circ, 100^\circ]$, $[-150^\circ, 0^\circ]$ and $[-45^\circ, 45^\circ]$. The state is $s = (q, \dot q) \in \mathbb{R}^{18}$, where $q = (z, p_x, \phi, q_{1:6})$ includes the torso's vertical and horizontal displacement and pitch, and the six leg joint angles. The action $a \in [-1, 1]^6$ is the normalized joint torque command, scaled by gear ratios of $100$, $50$ and $20$ at the hip, knee and ankle. Action knots are spaced $\Delta t = 0.15\,\mathrm{s}$ apart ($15$ simulator substeps), with the command held constant in between.

\paragraph{Cost.}
With the torso height $h = 1.3\,\mathrm{m} + z$ and the horizontal velocity
$v_{\mathrm{com}}$ of the whole-body center of mass, the per-state cost is
\begin{equation}
c(s) = w_h\,(h - h^*)^2 + w_{\mathrm{up}}\,(\cos\phi - 1)^2
        + w_v\,(v_{\mathrm{com}} - v^*)^2 ,
\end{equation}
and the planning objective over a horizon of $H$ knots is
\begin{equation}
J(s_{0:H}, a_{0:H-1}) = \Delta t \sum_{t=0}^{H-1}\big(c(s_t) + w_a \|a_t\|^2\big)
                        + c(s_H),
\end{equation}
with the targets $h^* = 1.2\,\mathrm{m}$ and $v^* = 1.5\,\mathrm{m/s}$ and the weights $w_h = 10$, $w_{\mathrm{up}} = 3$, $w_v = 1$, and $w_a = 0.1$, following Hydrax \citep{kurtz2024hydrax}. The upright term vanishes when the torso is vertical, and the center-of-mass velocity is computed from the state in closed form.

\paragraph{Problem generation.}
A problem instance corresponds to an initial state, where the root slides are zero (torso at its home
height), the pitch is uniform in $[-1.5, 1.5]\,\mathrm{rad}$, each leg joint is
uniform within its range, and every generalized velocity is uniform in
$[-0.1, 0.1]$. Since a pitch of $1.5\,\mathrm{rad}$ is nearly horizontal and
the legs are placed at random, many starts are fallen or falling, and the
planner has to stand the robot up before it can walk. The held-out evaluation
set consists of $64$ such starts drawn with a fixed seed and shared by all
methods.

\paragraph{Closed-loop evaluation.}
All planners run in receding horizon with $H = 4$ knots ($0.6\,\mathrm{s}$ of look-ahead) and re-plan at $50\,\mathrm{Hz}$, i.e., every two simulator steps. As in Hydrax, the knots are not shifted between re-plans, so the planner is warm-started from its previous plan and re-optimizes it from the new state. The executed action is the planner's nominal. The reported closed-loop cost is the mean running cost $c(s_t) + w_a\|a_t\|^2$ over the executed steps of the episode of $250$ closed-loop steps ($5\,\mathrm{s}$).

\subsubsection{Cheetah-WM environment}
\label{app:cheetah_wm}

Cheetah-WM is the \texttt{cheetah-run} task of the DeepMind Control Suite, solved by planning inside the latent world model of a pretrained TD-MPC2 agent \citep{hansen2024tdmpc2} instead of the physics simulator. All planners optimize the objective of TD-MPC2's own \mppi~planner, which isolates the optimizer from the model and replaces the hand-written cost with a neural network.

\paragraph{World model.}
We use the frozen single-task TD-MPC2 checkpoint for \texttt{cheetah-run} ($5$M parameters, latent dimension $512$, five Q-heads). The encoder maps the $17$-dimensional proprioceptive observation $o_t$ to a latent $z_t \in \mathbb{R}^{512}$, with deterministic latent dynamics $z_{t+1} = f(z_t, a_t)$. The action $a_t \in [-1, 1]^6$ is the torque command, clamped to the box inside the rollout and the cost, which matches the action range used to train the model. The reward head $R(z, a)$ and the Q-ensemble $Q_k(z, a)$ are TD-MPC2's two-hot decoded heads, and $\pi(z)$ is its policy prior.

\paragraph{Cost.}
For a latent $z_0$ and an action sequence $a_{0:H-1}$, the planning objective
is minus TD-MPC2's value estimate of the sequence,
\begin{equation}
J(a_{0:H-1};\, z_0) = -\Big[ \sum_{t=0}^{H-1} \gamma^t R(z_t, a_t)
  + \gamma^H\, \bar Q\big(z_H, \pi_{\mathrm{mean}}(z_H)\big) \Big],
\qquad z_{t+1} = f(z_t, a_t),
\end{equation}
with the horizon $H = 3$ and discount $\gamma$ of the checkpoint, $\pi_{\mathrm{mean}}$ the mean of the policy prior at the terminal latent, and $\bar Q$ the mean over the five Q-heads. TD-MPC2's planner instead averages a random two-head subsample at every evaluation, whereas the deterministic mean gives every candidate the same objective. Termination heads are ignored, as the task is non-episodic.

\paragraph{Problem generation.}
A problem instance corresponds to an initial latent, obtained by encoding the initial observation of a \texttt{dm\_control} episode. Episode $k$ of planner seed $s$ is reset with seed $1000\,s + k$, so that all methods share the same start states, and we run $3$ planner seeds of $10$ episodes each ($30$ paired instances). Planner settings are chosen on a separate imagined closed loop of $64$ held-out latents rolled out for $500$ steps inside the frozen model, and these numbers are not reported.

\paragraph{Closed-loop evaluation.}
At every environment step, the observation is encoded, the planner optimizes $J$ over $H = 3$ actions from the current latent, and the first action is applied to the real environment. The next step is warm-started from the plan shifted by one action with a zero action appended. Since TD-MPC2 itself executes an elite sample rather than the population mean, the sampling-based planners execute the cheapest sample of their final population. The reported metric is the undiscounted return of the real environment over an episode of $500$ decision steps (the sum of the true task rewards, in $[0, 1000]$), normalized by the mean return of the best method.

\subsubsection{Humanoid-WM environment}
\label{app:humanoid_wm}

Humanoid-WM is the \texttt{humanoid-walk} task of the DeepMind Control Suite, solved by planning inside the latent world model of a pretrained TD-MPC2 agent \citep{hansen2024tdmpc2}, with the same construction as Cheetah-WM. This is the hardest of our tasks, as the humanoid has $21$ actuators and must keep its balance while walking, so a poor short-horizon plan makes it fall within a few steps. The gap between low- and high-budget sampling-based planners is thus the largest here, making this the clearest test of whether \flowoptpolicy matches the high-budget teacher at a small budget.

\paragraph{World model.}
We use the frozen single-task TD-MPC2 checkpoint for \texttt{humanoid-walk} ($5$M parameters, latent dimension $512$, five Q-heads). The setup is identical to Cheetah-WM, except that the proprioceptive observation is $67$-dimensional and the action is $a_t \in [-1, 1]^{21}$.

\paragraph{Cost, problem generation and closed-loop evaluation.}
The planning objective, the generation of problem instances ($3$ planner seeds of $10$ episodes each, $30$ paired instances) and the closed-loop evaluation are identical to those of Cheetah-WM (Appendix~\ref{app:cheetah_wm}).

\begin{table}[t]
\centering
\caption{Average performance metrics on six main tasks presented in Section \ref{sec: experiments - comp with sampling based}.}
\label{tab: main - metrics}
\vspace{0.15cm}
\small
\setlength{\tabcolsep}{5pt}
\renewcommand{\arraystretch}{1.0}
\newcommand{\taskm}[2]{\begin{tabular}[c]{@{}l@{}}#1\\[-3pt]{\scriptsize (#2)}\end{tabular}}
\begin{tabular}{@{}l ccccc c@{}}
\toprule
Task & \mppi & \cem & \cmaes & \dialmpc & \lgu & \textbf{\flowoptpolicy (Ours)} \\
\midrule
\taskm{CarNav}{$J_{\mathrm{norm}}\!\downarrow$}      & \ms{1.19}{0.40} & \ms{1.17}{0.41} & \ms{1.13}{0.39} & \ms{1.38}{0.49} & \ms{1.06}{0.37} & \textbf{\ms{1.00}{0.33}} \\
\addlinespace[3pt]
\taskm{Cheetah}{$R_{\mathrm{norm}}\!\uparrow$}       & \ms{0.66}{0.05} & \ms{0.61}{0.05} & \ms{0.71}{0.05} & \ms{0.58}{0.04} & \ms{0.90}{0.07} & \textbf{\ms{1.00}{0.09}} \\
\addlinespace[3pt]
\taskm{Walker}{IQM $J_{\mathrm{norm}}\!\downarrow$}        & 2.53 & 1.38 & 1.69 & 1.44 & 1.07 & \textbf{1.00} \\
\addlinespace[3pt]
\taskm{Push-T}{SR\,\%\,$\uparrow$}                   & \ms{60.4}{4.8}  & \ms{38.0}{2.4}  & \ms{54.7}{5.6}  & \ms{53.1}{4.1}  & \ms{86.5}{11.0} & \textbf{\ms{99.0}{0.9}} \\
\addlinespace[6pt]
\taskm{Cheetah-WM}{$R_{\mathrm{norm}}\!\uparrow$}    & \ms{0.86}{0.06} & \ms{0.88}{0.07} & \ms{0.87}{0.04} & \ms{0.79}{0.04} & \ms{0.95}{0.16} & \textbf{\ms{1.00}{0.09}} \\
\addlinespace[3pt]
\taskm{Humanoid-WM}{$R_{\mathrm{norm}}\!\uparrow$}   & \ms{0.39}{0.10} & \ms{0.42}{0.11} & \ms{0.35}{0.08} & \ms{0.53}{0.16} & \ms{0.88}{0.12} & \textbf{\ms{1.00}{0.08}} \\
\bottomrule
\end{tabular}
\end{table}

\begin{table}[t]
\centering
\caption{CarNav: closed-loop cost (mean $\pm$ SE over 100 paired problems) on the training and shifted settings, and number of problems on which \flowoptpolicy~is better.}
\label{tab:carnav_ood}
\vspace{0.15cm}
\begin{tabular}{lrrr}
\toprule
Setting & Ours & \flowpolicy & Ours better \\
\midrule
Training & $7095 \pm 163$ & $7639 \pm 375$ & $73/100$ \\
Wheelbase 2 & $7329 \pm 182$ & $9132 \pm 497$ & $84/100$  \\
4 obstacles & $7876 \pm 290$ & $10675 \pm 597$ & $86/100$ \\
8 obstacles & $8366 \pm 401$ & $11632 \pm 538$ & $93/100$ \\
16 obstacles & $8878 \pm 534$ & $13239 \pm 626$ & $97/100$ \\
\bottomrule
\end{tabular}
\end{table}

\begin{table}[t]
\centering
\caption{Push-T: mean stage cost (mean $\pm$ SE over 64 paired problems), number of problems on which \flowoptpolicy is better, and success rate on the training and shifted settings.}
\label{tab:pusht_ood}
\vspace{0.15cm}
\begin{tabular}{lrrrrr}
\toprule
 & \multicolumn{2}{c}{Cost} & & \multicolumn{2}{c}{Success (\%)} \\
\cmidrule(lr){2-3} \cmidrule(lr){5-6}
Setting & Ours & \flowpolicy & Ours better & Ours & \flowpolicy \\
\midrule
Training & $0.280 \pm 0.035$ & $0.359 \pm 0.045$ & 63/64 & 100.0 & 59.4 \\
Friction 0.5 & $0.348 \pm 0.044$ & $0.455 \pm 0.059$ & 59/64 & 92.2 & 51.6 \\
Friction 2 & $0.242 \pm 0.031$ & $0.287 \pm 0.036$ & 56/64 & 100.0 & 43.8 \\
Start range 0.4 & $0.447 \pm 0.055$ & $0.532 \pm 0.062$ & 59/64 & 96.9 & 39.1 \\
\bottomrule
\end{tabular}
\end{table}

\begin{table}[t]
\centering
\caption{Push-T with an additional control-rate penalty $w_{\Delta u}$, absent during training. The model-based planners optimize the modified objective, while \flowpolicy~is evaluated without retraining. \mppi~and \cem~retain the hyperparameters tuned on the original objective. We report the mean closed-loop stage cost under the modified objective, the control-rate term and the success rate.}
\label{tab:objective_shift}
\vspace{0.15cm}
\small
\begin{tabular}{lccccc}
\toprule
 & \multicolumn{5}{c}{$w_{\Delta u}$} \\
\cmidrule(lr){2-6}
 & $0$ (train) & $3 \times 10^{-3}$ & $1 \times 10^{-2}$ & $3 \times 10^{-2}$ & $1 \times 10^{-1}$ \\
\midrule
\multicolumn{6}{l}{\textbf{Cost under modified objective} $\downarrow$} \\
Reference (high budget) & 0.29 & 0.31 & 0.34 & 0.41 & 0.47 \\
\flowoptpolicy          & 0.28 & 0.35 & 0.36 & 0.43 & 0.51 \\
\mppi~(equal budget)     & 0.36 & 0.37 & 0.43 & 0.54 & 0.67 \\
\cem~(equal budget)      & 0.38 & 0.39 & 0.44 & 0.52 & 0.68 \\
\flowpolicy & 0.36 & 0.38 & 0.41 & 0.51 & 0.92 \\
\midrule
\multicolumn{6}{l}{\textbf{Control-rate term} $\downarrow$} \\
Reference (high budget) & 7.2 & 6.4 & 5.1 & 3.1 & 1.2 \\
\flowoptpolicy          & 7.1 & 7.2 & 5.7 & 3.5 & 1.4 \\
\mppi~(equal budget)     & 8.0 & 8.4 & 6.4 & 4.3 & 2.5 \\
\cem~(equal budget)      & 8.8 & 7.9 & 7.7 & 4.6 & 2.7 \\
\flowpolicy & 5.3 & 5.3 & 5.3 & 5.3 & 5.3 \\
\midrule
\multicolumn{6}{l}{\textbf{Success (\%)} $\uparrow$} \\
Reference (high budget) & 99 & 98 & 93 & 86 & 79 \\
\flowoptpolicy          & 99 & 88 & 85 & 83 & 71 \\
\mppi~(equal budget)     & 60 & 54 & 56 & 42 & 25 \\
\cem~(equal budget)      & 38 & 39 & 35 & 25 & 12 \\
\flowpolicy & 59 & 59 & 59 & 59 & 59 \\
\bottomrule
\end{tabular}
\end{table}

\subsection{Planning Details}

\paragraph{Receding-horizon loop.}
Every method plans a sequence of $H$ control knots at each closed-loop step, executes the first knot, and is warm-started at the next step from its previous plan. All tasks use $K = 3$ optimization iterations and a population size of $P = 30$, except Humanoid-WM, where ${P = 150}$. In \flowoptpolicy, the velocity field is integrated with $5$ Euler steps, and the hyperparameters of the baselines are tuned by grid search on a separate validation set.

\paragraph{Teacher optimizer.}
The teacher is \mppi, run once per member of the current population $\vx^k$. Each search is initialized at that member with a fixed per-control standard deviation of $0.3\,\sigma_{\mathrm{init}}$ and uses $K_T$ iterations of $P_T$ rollouts, with $(P_T, K_T) = (100, 20)$ on all tasks except Humanoid-WM, where $(P_T, K_T) = (500, 20)$. The field is refitted every $N_E = 64$ episodes, with $\tilde{\lambda} = 0.125$ and $\tilde{h} = 1$.

\paragraph{Learned Gaussian Update (LGU).}
This baseline isolates the effect of transporting candidates individually from that of learning how to update the parameters of a Gaussian proposal distribution. The method maintains a diagonal Gaussian proposal $\mathcal{N}(\mu^k, \mathrm{diag}(\sigma^k)^2)$ over action sequences. At iteration $k$, it draws $P$ candidates from the proposal, evaluates them, and a network maps the evaluated population and its context $C^k$ to the updated parameters $(\mu^{k+1}, \sigma^{k+1})$, from which the next population is drawn, in the spirit of \citet{sacks2022learning}. All remaining components are shared with \flowoptpolicy. \lgu~is trained on the same closed-loop data and with the same teacher, by maximizing the Gaussian likelihood of the target population $\vx^{k,+}$ onto which \flowoptpolicy is regressed. It uses the same $P$ and $K$, hence the same total of $(K+1)P$ rollouts per control step.

\paragraph{Flow Policy.}
This baseline serves to compare optimizing through a predictive model with generating actions directly from the state. The policy is a conditional flow model that maps Gaussian noise to action sequences given the current state $\hat{\vs}_0$ and, where applicable, task information such as the goal and the obstacle positions. The network uses the velocity head of \flowoptpolicy (Appendix~\ref{app:head}), with the attended token replaced by an embedding of the state and task information, so that neither the population nor its costs are observed. On CarNav, the obstacles enter through a fixed number of input slots, padded or truncated when their number differs from that of training. \flowpolicy~is trained with the same procedure and teacher as \flowoptpolicy, by flow matching onto the same target populations $\vx^{k,+}$. At deployment, \flowpolicy~generates $P$ action sequences per control step, the same number of candidates as \flowoptpolicy, and executes the best of them.

\subsection{Training Details}

The architecture of the velocity field is described in Appendix~\ref{app:architecture}, with $d = 192$, $n_h = 4$ and $m = 512$. We train with Adam at a learning rate of $10^{-4}$, a batch size of $64$ populations and gradient-norm clipping at $1$.

\subsection{Additional Results}

Table \ref{tab: main - metrics} provides the average metrics across all tasks for every method accompanying Section \ref{sec: experiments - comp with sampling based}. 
The table reports cost for CarNav and Walker (lower is better), return for HalfCheetah, Cheetah-WM and Humanoid-WM (higher is better), and success rate for Push-T. Costs and returns are normalized by the mean of the best method on each task, so that the best method scores $1$. For Walker, we report the interquartile mean of the normalized cost.

The performance profiles of Figure~\ref{fig: performance profiles} are computed per evaluation episode \citep{dolan2002benchmarking}. For each episode, the cost of every method is divided by the lowest cost attained on that episode, and the profile of a method at $\tau$ is the fraction of episodes on which this ratio is at most $\tau$. The per-episode cost is the closed-loop cost on CarNav and Walker, the mean stage cost on Push-T, and the negated return $J = -R$ on HalfCheetah, Cheetah-WM and Humanoid-WM.

Tables~\ref{tab:carnav_ood} and \ref{tab:pusht_ood} report the full generalization results of Section~\ref{sec: experiments - generalization}, comparing \flowoptpolicy with \flowpolicy~on the training distribution and on shifted settings that alter the environment (number of obstacles on CarNav), the dynamics (wheelbase on CarNav, friction on Push-T), the initial-state distribution (start range on Push-T), or the objective (an added control-rate penalty $w_{\Delta u}$ on Push-T).

Table~\ref{tab:objective_shift} further evaluates generalization to an unseen objective by introducing a control-rate penalty on Push-T at deployment, without retraining. As the penalty increases, \flowoptpolicy adapts its actions, reducing the unweighted control-rate term from $7.1$ to $1.4$, whereas \flowpolicy~remains unchanged at $5.3$. Even at the largest penalty, \flowoptpolicy retains a $71\%$ success rate, compared with $25\%$ for equal-budget \mppi~and $12\%$ for \cem, approaching the $79\%$ achieved by the higher-budget reference. These results demonstrate the ability of our learned optimizer to adapt to changes in the objective through rollout feedback alone.

\begin{figure}[t]
    \centering
    \includegraphics[width=0.5\textwidth]{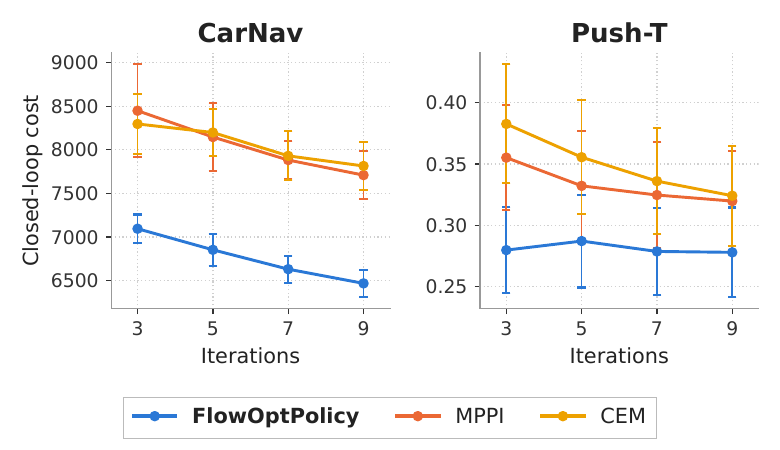}
    \caption{The effect of increasing iterations in the CarNav and Push-T tasks.}
    \label{fig:ablation}
\end{figure}

Figure~\ref{fig:ablation} accompanies the ablations of Section~\ref{sec: experiments - ablations}, showing the effect of increasing the number of optimization iterations per control step beyond the training budget on CarNav and Push-T.

\paragraph{Ablation variants (Table~\ref{tab:component_ablations}).}
All variants are evaluated with the deployment budget, evaluation instances and seeds of the full model. The \emph{Flow integration} variants reuse the trained field of the full model and change only the number of Euler steps at deployment, $N_\tau \in \{1, 2, 10\}$ in place of the default $5$. All other variants are retrained with the same data collection and training procedure.
\emph{No attention:} the attention output is removed, $a_i = \mathrm{LN}_2(q_i)$, so each member is updated from its own token and the flow time only. The other members are observed only through population statistics, namely the standardization of its plan and its own cost rank and standardized cost.
\emph{Z-score only} and \emph{Rank only:} the context of each member is reduced to $\tilde{J}_j$ and to $\rho_j$, respectively.
\emph{Teacher quality:} the teacher's rollout budget $K_T P_T$ is divided by $4$ and by $16$, by lowering both the number of iterations $K_T$ and the rollouts per iteration $P_T$.
\emph{Network size:} $(d, m) \in \{(64, 256), (384, 1024)\}$ against the default $(192, 512)$, with $d$ and $m$ as in Appendices~\ref{app:blocks} and~\ref{app:head}.
Differences are computed with respect to the full model on the same instances, with $\pm$ the standard error of the paired difference.

\end{document}